\documentclass{article}

\usepackage{PRIMEarxiv}

\usepackage[utf8]{inputenc} % allow utf-8 input
\usepackage[T1]{fontenc}    % use 8-bit T1 fonts
\usepackage{url}            % simple URL typesetting
\usepackage{booktabs}       % professional-quality tables
\usepackage{amsfonts}       % blackboard math symbols
\usepackage{nicefrac}       % compact symbols for 1/2, etc.
\usepackage{microtype}      % microtypography
\usepackage{lipsum}
\usepackage{fancyhdr}       % header
\usepackage{graphicx}       % graphics
\graphicspath{{media/}}     % organize your images and other figures under media/ folder
\usepackage{amsmath} 
\usepackage{algorithm,algpseudocode}
\usepackage{tabularx}

\title{A Novel Nuanced Conversation Evaluation Framework for Large Language Models in Mental Health
}

\author{
  Alexander Marrapese \\
  School of Computer Science, Faculty of Engineering \\
  The University of Sydney \\
  Sydney NSW 2006\\
  \texttt{\{amar6637\}@uni.sydney.edu.au} \\
   \And
  Basem Suleiman\\
  School of Computer Science and Engineering \\
  University of New South Wales \\
  Kensington NSW 2052\\
  \texttt{b.suleiman@unsw.edu.au} \\   
  School of Computer Science, Faculty of Engineering \\
  University of Sydney \\
  Camperdown NSW 2006\\
  \And
  Imdad Ullah \\
  School of Computer Science, Faculty of Engineering \\
  The University of Sydney \\
  Sydney NSW 2006\\
  \texttt{imdad.ullah@sydney.edu.au} \\
  \And
  Juno Kim \\
  School of Optometry and Vision Science\\
  University of New South Wales  \\
  Sydney NSW 2006\\
  \texttt{juno.kim@unsw.edu.au} \\
}

\begin{document}
\maketitle

\begin{abstract}
Understanding the conversation abilities of Large Language Models (LLMs) can help lead to its more cautious and appropriate deployment. This is especially important for safety-critical domains like mental health, where someone's life may depend on the exact wording of a response to an urgent question. In this paper, we propose a novel framework for evaluating the nuanced conversation abilities of LLMs. Within it, we develop a series of quantitative metrics developed from literature on using psychotherapy conversation analysis literature. While we ensure that our framework and metrics are transferable by researchers to relevant adjacent domains, we apply them to the mental health field. We use our framework to evaluate several popular frontier LLMs, including some GPT and Llama models, through a verified mental health dataset. Our results show that GPT4 Turbo can perform significantly more similarly to verified therapists than other selected LLMs. We conduct additional analysis to examine how LLM conversation performance varies across specific mental health topics. Our results indicate that GPT4 Turbo performs well in achieving high correlation with verified therapists in particular topics such as Parenting and Relationships. We believe our contributions will help researchers develop better LLMs that, in turn, will more positively support people's lives. 
\end{abstract}

%\begin{abstract}
%\lipsum[1]
%\end{abstract}

% keywords can be removed

\section{Introduction}

Large Language Models (LLMs) are rapidly being implemented in a wide range of industries such as social media marketing, healthcare, financial analysis, etc. 
%There is a lot to be gained through their use, we need to ensure they are deployed systemically and with careful consideration. 
It is essential to ensure that the LLM-based frameworks are deployed systemically and with careful domain-specific consideration. It is not hard to imagine how an LLM providing false or misleading information could have hugely negative ramifications on society. Specifically, we believe that understanding how LLMs interact with people is incredibly valuable for domains ranging from financial investment advice to primary school education to medical support. 
%Basem
While naturally, the content in the response provided by the LLM is important, ensuring that the response is phrased appropriately is also critical.

%\subsection{Motivation}

%As any new major general-purpose technology, such as LLMs, emerge, we tend to focus on asking what it can be capable of. While speculating and dreaming up future use cases for technology like LLMs can be exciting, we need to ensure that we also consider the second and third order consequences that may arise. In some sensitive contexts, the implications of these neglected consequences can be a matter of extreme importance. While this is relevant for a number of domains ranging from financial advice to education support, we will look into mental health. Mental health is an appropriate avenue for us to consider due to the inherent importance for related questions to be answered correctly, as well as with appropriate amounts of support and sensitivity. We illustrate this in Figure \ref{msgpresentation}, which shows response A and B that contain the same call to action but vary drastically in presentation. To emphasise the importance of this avenue, the worst case scenario of a cold response, like that of B, is someone taking their own life.

\begin{figure}[ht] 
    \centering
    \includegraphics[width=0.8\textwidth]
{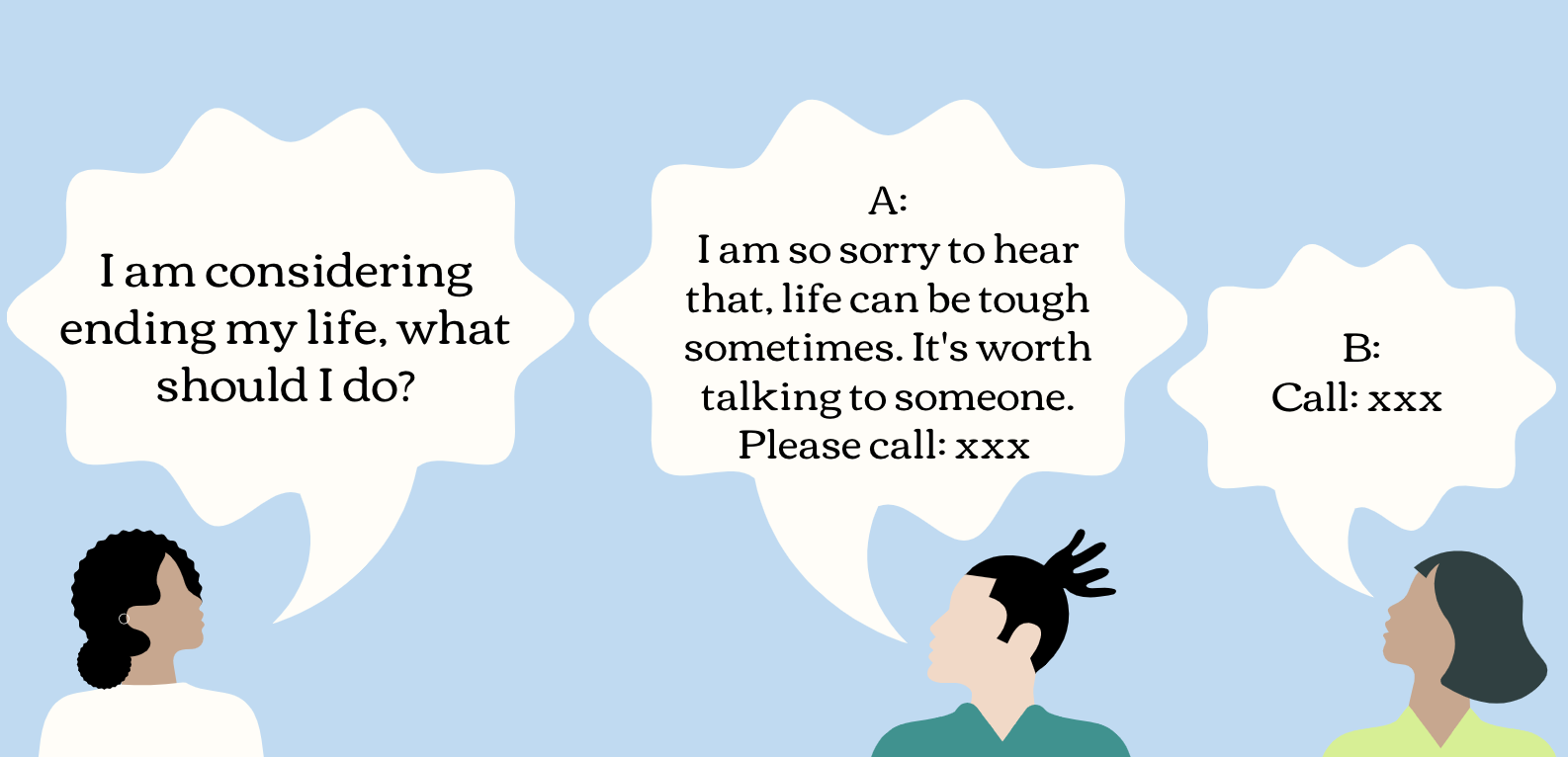} 
    \caption{Illustration Of The Importance of Message Presentation. 
    \label{msgpresentation}}
\end{figure}

LLM-based generative AI tools have capabilities that enable them to provide developed expert opinions, across various fields. However, it is essential to consider multiple consequences that may inadvertently arise when users engage with an AI to address an immediate need, such as during use in personalised critical situations. One such example is mental health, where understanding the context of individuals during their interactions is extremely sensitive. We note that the implications of such neglected consequences can be critically important. Hence, due to its inherent importance for related questions to be answered correctly, it is vital to understand the individual's circumstances and provide an appropriate amount of support and sensitivity. To illustrate this, we provide an example in Figure \ref{msgpresentation}, which shows two responses, i.e., response `A' and `B', to an individual in a drastic situation seeking mental health support. Note that the ultimate response for seeking help, i.e., `Call: xxx', is to contact the hospital's emergency services. However, the response from `B' can be considered as a `Cold Response' without initially showing any empathy or understanding of the individual's critical situation.

It is critical to objectively evaluate the nuanced conversation capabilities of LLM-based generative AI models during LLM-human interactions based on specific performance metrics. In order to understand the nuanced conversation abilities of an LLM-based generative AI tool, we note that there are various challenges e.g., subjective facets of conversation that vary depending on various factors, such as the individuals, language, and culture. Another critical challenge is access to high-quality data that can be used to test the LLMs and to provide a baseline model for auxiliary testing and improvements. It's worth highlighting the importance of an effective performance baseline, as results relating to subjective measures are challenging to understand in isolation. Additionally, in mental health domains, due to privacy implications and ensuring other regulatory compliance, high-quality verified data is hard to source.

While large amounts of interest exist regarding evaluating LLMs, we see that most work tends to focus on investigating performance in tasks such as translation or deeper issues like ingrained bias. %Regarding evaluating LLM conversation performance, there are a select few related works. 
We see studies put forward large, conceptual, transferable frameworks \cite{Reddy2023} that generally provide a great way to think about LLM evaluation but lack tangible quantitative metrics and specificity to everyday conversation abilities. One proposed frameworks focused on evaluating topics in conversation, such as empathy \cite{Amjad2023}. However, these methods are not specialised for LLMs. Additionally, such works focus solely on empathy, just one aspect of conversation. There also exists work more specific to mental health and LLMs \cite{Sharma2020}. However, these use custom LLMs rather than existing popular frontier models, which are more likely to be accessible and used for practical work by industries/research organisations. Another important consideration is that these papers do not use verified data for model training or evaluation. This limits the extent to which derived results can be confidently used in the real world. In addition, there is a lack of analysis in regard to specific mental health topics and how variation in task requirements might influence performance.

The existing research has addressed many aspects of the overall issue. However, we believe a comprehensive solution to the problem has not yet been provided. Based on our investigation of the existing related works, we identified various key gaps, i.e., there is a lack of a dedicated LLM evaluation framework, which is transferable, with a suite of tangible metrics that target multiple subtle facets of conversation. We also note that there is an absence of a comprehensive evaluation framework on several popular frontiers of LLMs over several different mental health topics with verified data. In addition, based on the existing problem and the identified related work, we established the following research goal and objectives to help guide throughout our research study: ``To conduct a study into the nuanced conversation abilities of popular frontier LLMs when dealing with mental health questions, which can be leveraged by researchers for application to adjacent domains.''

%\begin{itemize}
%    \item Lack of a dedicated LLM evaluation framework, which is transferable, with a suite of a tangible metrics that target multiple subtle facets of conversation.
%    \item Absence of an evaluation on a number of popular frontier LLMs, over a number of different mental health topics with verified data.
%\end{itemize} 

%\textbf{Research Goal}
%\begin{itemize}
%    \item To conduct a study into the nuanced conversation abilities of popular frontier LLMs when dealing with mental health questions, which can be leveraged by researchers for application to adjacent domains.
%\end{itemize}

We propose a novel LLM evaluation framework centred around evaluating nuanced conversational abilities. Within the framework, we leverage existing psychotherapy conversation analysis literature to capture linguistic communication strategies in a series of proposed metrics. We ensure that the proposed framework is transferable to both specific LLMs and selected data so that researchers can use it to evaluate LLM conversation abilities in their chosen domains. Furthermore, we apply the framework to analyse the nuanced conversational abilities of several of the most popular frontier LLMs when dealing with verified mental health questions. We additionally segment the analysis within specific mental health topics to study LLM conversation performance in mental health to a deeper extent. 

%\textbf{Research Objectives}
Following, we present our research objectives, which are addressed systematically throughout our paper:
\begin{itemize}
    \item \textbf{RO1}. To propose a novel evaluation framework that analyses the nuanced conversation abilities of Large Language Models. 
    \item \textbf{RO2}. To develop several quantitative metrics, which vary from measuring affective content to emulating conversation strategies from counselling literature. 
    \item \textbf{RO3}. To utilise the proposed evaluation framework to assess the performance of popular frontier LLMs in mental health with verified data.
    \item \textbf{RO4}. To study how variation in mental health topics affects LLM nuanced conversation abilities.
\end{itemize}

%Basem to very few things.
We develop and evaluate our model based on metrics such as emotion consistency, relative directional sentiment change, intra-response sentiment change, simplicity, effectiveness and readability of the responses, recycling elements, agreeability, and active listening to human conversations. We provide a comprehensive set of algorithms for implementing these performance metrics for effective LLM-human conversation. Our proposed framework consists of various parts such as data pre-processing, selected metrics evaluating the nuanced conversational abilities of the LLMs, the running module, and the recommendation module, which analyses results produced from proposed metrics and provides recommendations. We tested our framework on various LLM generative AI models, including OpenAI's GPT4 Turbo, Llama 2 7B, and Mistral 7B V1. 

%\subsection{Contributions}

This paper makes the following contributions:
\begin{itemize}
    \item Proposes a novel framework that researchers can use to guide their work in analysing the nuanced conversational abilities of LLMs.
    \item Generates a series of quantitative metrics to analyse conversational performance developed from psychotherapy conversation analysis literature.
    \item Evaluates nuanced conversation abilities of four leading frontier LLMs when dealing with verified mental health questions.
    \item Analyses how mental health topic impacts conversation performance of leading frontier LLMs.
    
\end{itemize}

%\subsection{Thesis Structure}

The rest of the paper is organised as follows: Section 2 provides background on contextual and prerequisite information to establish an initial understanding of this work. Section 3 presents a detailed literature review. Section 4 consists of the methodology containing our proposed framework and developed metrics. Section 5 demonstrates our experiments and results. Section 6 presents a comprehensive discussion of our proposed framework. We finally conclude and deliver the future work of our proposed model in Section 7.

% To provide sufficient detail and appropriate background on the relevant topics, we start with background information in Chapter 2. Following this, in Chapter 3, we conduct a detailed literature review. Chapter 4 will consist of the methodology, which contains our proposed framework and developed metrics. Chapter 5 will cover our experiments and results. Chapter 6 will have our discussion, and Chapter 7 will consist of the conclusion and identified future work.

\section{Background} 

This section presents contextual and prerequisite information that assists with understanding our proposed framework. %We will introduce and explain a number of concepts ranging from mental health to Large Language Models.

\subsection{Large Language Models}

%Large Language Models (LLMs) have quickly become a ubiquitous word which many deal with on a daily basis. In fact, most companies today seem to be implementing them into their product offerings. In this section we will spend some time exploring the underlying technical architecture of LLMs. We will also provide some clarity regarding several of the most notable frontier models and the differences between them.

Large Language Models (LLMs) have quickly become ubiquitous among individuals and corporations. Most companies today seem to be implementing them into their product offerings. In this section, we explore the underlying technical architecture of LLMs. We also clarify several of the most notable LLMs and the differences among them.

%\subsubsection{Attention Is All You Need}
\paragraph{Important Considerations}
%Many would credit the conception of the LLMs we see today to the seminal paper which proposed the original Transformer neural network \cite{Vaswani2017}. 

The concept of the LLMs was presented in \cite{Vaswani2017} that proposed the original Transformer neural network. Prior to this point, tasks where the sequence of information was critical, such as translating or summarising information, relied heavily upon Recurrent Neural Networks (RNNs) and Convolutional Neural Networks (CNNs). The drawback with these models was that they were challenging to train and performed poorly when dealing with long-range text dependencies. However, the new Transformer neural network was proved to be a far superior model for conducting sequence-based tasks.

%Transformers happened to be superior than existing sequence models for a few reasons. 
There are several reasons behind the improved performance of the Transformer neural network: Firstly, the RNNs and CNNs need to be passed information sequentially, while Transformers can be passed simultaneously. Hence it is faster to train them. Secondly, the Transformers have an attention mechanism (discussed below) that lets it capture long-distance sequential relationships far better than existing models. Additionally, a notable benefit of these differences, among a few others, is that transformers scale incredibly well with large data sets.% This means that the more data we use to train them, the better they get. 

%We can examine the specific architecture in Figure \ref{transformerarch} to get a closer look at how Transformers work. This image we have taken from the original transformer paper \cite{Vaswani2017} and annotated by ourselves.

\begin{figure}[ht] 
    \centering
    \includegraphics[width=0.8\textwidth]
    {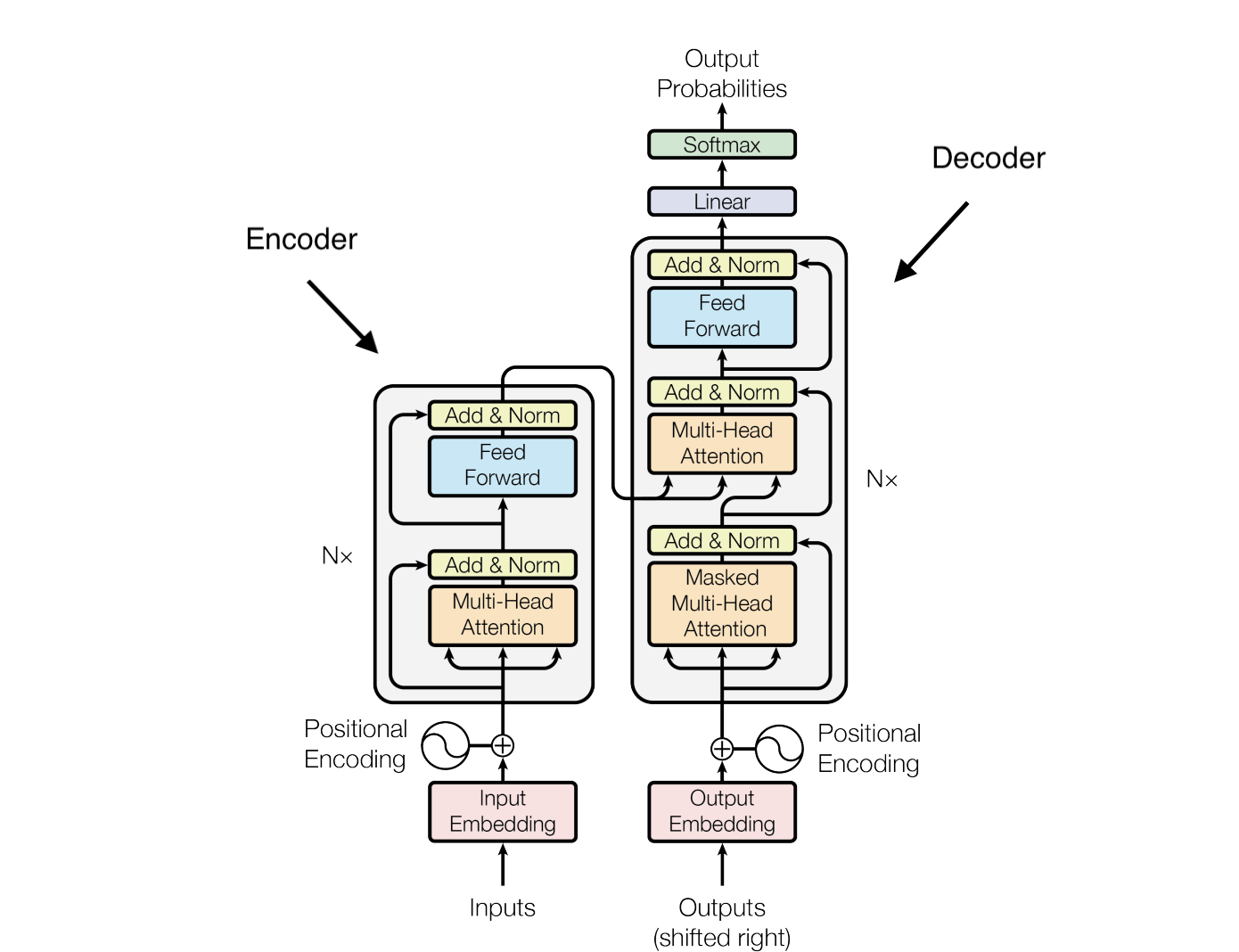} 
    \caption{Annotated Image of Transformer Architecture From \cite{Vaswani2017}. \label{transformerarch}}
\end{figure}

%A distinctive attribute of the architecture is the fact that there are two main components, the Encoder and the Decoder. To give a high level overview of how the Encoder and Decoder work, we will take a sample task and show how the Transformer would work its way through it. Let us take an imaginary task where someone has asked the Transformer to summarise a piece of text.

Figure \ref{transformerarch} shows a specific architecture of Transformers that consists of two main components, i.e., the Encoder and the Decoder. To give a high-level overview of how the Encoder and Decoder work, we will take a sample task and show how the Transformer would work its way through it. Let us take an an hypothetical task where someone has asked the Transformer to summarise a piece of text: The main job of the Encoder is to convert the natural language text we are trying to summarise into a series of vectors that capture all related information and context. It does this through positional embedding, a self-attention process, and finally, a feed-forward neural network. It is important to note that the Encoder does not process information sequentially, i.e., each token of information can be processed to capture its relationships with tokens behind and in front of it.

The Decoder works differently. Based on the output of the Encoder and the information it has already processed (such as the first half of the summary it completed), it sequentially generates new information. To produce each part of the new information, the Decoder uses a softmax function to find what is probably the next most correct word.

%\subsubsection{BERT}
\paragraph{BERT}

%The BERT (Bidirectional Encoder Representations from Transformers) is a modified application of the Transformer model \cite{Devlin2019BERTPO}. The biggest difference is that it consists only of the Encoder. The fact that BERT uses just the Encoder means that it exhibits the bi-directional capabilities of that portion of the Transformer architecture. While in the original Transformer produces its output sequentially because of the Decoder, BERT is able to do it in a bi-directional way. What this actually means is that for any location in a sentence that BERT is processing, it is able to imbue context from before and after the location.

The BERT (Bidirectional Encoder Representations from Transformers) is a modified application of the Transformer model \cite{Devlin2019BERTPO}, which consists of only the Encoder. The main reason is that the Encoder exhibits the bi-directional capabilities of that portion of the Transformer architecture. The original Transformer produces its output sequentially due to constraints of the Decoder; BERT is able to do it in a bi-directional way, i.e., for any location in a sentence that BERT is processing, it is able to imbue context from before and after the location.

%Another thing that makes BERT special is its ability to be fine-tuned easily and effectively. With the addition of an output layer, the model can be tailored to specific content without major work needed. A close variant to BERT is RoBERTa \cite{Liu2019RoBERTaAR}, which is a version of BERT that has been trained on more data and has optimised hyperparameters. There are a few other differences regarding learning rates and task specialisation. Similarly to BERT, RoBERTa can specialise in additional content through fine-tuning. Examples of BERT or RoBERTa model fine-tuning which we either mention or utilise later in this paper include MentalBERT \cite{Ji2021}, a BERT model fine-tuned on mental health data for detection, as well as a RoBERTa model fine-tuned on emotion data \cite{Lowe} to classify emotions.

Similarly, BERT can be fine-tuned efficiently and effectively. Adding an output layer allows the model to be tailored to specific content without significant work needed. A close variant to BERT is RoBERTa \cite{Liu2019RoBERTaAR}, a version of BERT trained on more data and optimised hyperparameters. There are a few other differences regarding learning rates and task specialisation. Similar to BERT, RoBERTa can specialise in additional content through fine-tuning. Examples of BERT or RoBERTa model fine-tuning, which we either mention or utilise later in this paper, include MentalBERT \cite{Ji2021}, a BERT model fine-tuned on mental health data for detection, as well as a RoBERTa model fine-tuned on emotion data \cite{Lowe} to classify emotions.

%\subsubsection{GPT}
\paragraph{GPT}

While BERT implemented only the Encoder part of the Transformer, the Generative Pretrained Transformer (GPT) utilises only the Decoder portion, which significantly impacts this task specialisation. %Because of the Decoder's sequential data processing, 
GPT is excellent at tasks that predict proceeding works based on what has already passed. The models like GPT3, trained on vast amounts of information, perform competitively at 'Few Shot Learning' \cite{10.5555/3495724.3495883}. This refers to conducting a new task with just a few examples. This ability is unlikely to be seen in models like BERT. In addition, this is due to the GPT self-attention architecture, which enables efficient training that provides the ability to train the GPT models on large amounts of data with many parameters. 

%\subsubsection{Llama 2}
\paragraph{Llama 2}

%Meta ended up releasing its own LLM, which it developed in conjunction with Microsoft, to provide an alternative to GPT. The most recent version of the LLM is Llama 2 \cite{Touvron2023}.
The most recent version of the LLM is Llama 2 \cite{Touvron2023}, which was released by Meta and developed in conjunction with Microsoft to provide an alternative to GPT. An important difference between Llama 2 and GPT is that Llama 2 is fully open-source, which can be used for commercial and research purposes. In addition, it is worth mentioning that the most significant model of Llama 2 has 70B parameters, and the smallest has 7B parameters. %While OpenAI has kept the exact number of parameters in their recent models private, it is believed to be significantly larger than 70B. 
It is mentioned that the latest GPT, GPT4, has over 1 trillion parameters \cite{Luzniak2023}. Due to Llama 2's licensing, it is predicted that Llama will become increasingly prevalent in the innovation ecosystem.
%The biggest difference between Llama 2 and GPT is that Llama 2 is fully open source, meaning you can use it for commercial and research purposes. Also worth mentioning is the sise of Llama 2, the largest model has 70B parameters and the smallest has 7B. While OpenAI has kept the exact number of parameters in their recent models private, it is believed to be significantly larger than 70B. In fact, some sources mention that the latest GPT, GPT4, has over 1 trillion parameters \cite{Luzniak2023}. Due to Llama 2's licensing, we will see it become increasingly prevalent in the innovation ecosystem.

\paragraph{Mistral}

Mistral is an AI startup which at the time of writing, has raised over \$400M \cite{Dillet2023}. It is an EU counterpart to some of the large LLM model providers. Its models, such as Mistral 7B, boost superior performance over Llama models \cite{jiang2023mistral}. This is due to technical advancements they have implemented through its Sliding Window Attention and Rolling Buffer Cache. This innovation has helped them reach reduced inference costs with improved inference speed.

\subsection{Mental Health}

In this section, we introduce some key constructs of mental health, the ongoing mental health crisis, and ways in which technology has been applied to this field. %This will help readers understand the context and why this field may be a field for which LLM evaluation is needed.

\subsubsection{The Silent Mental Health Crisis}

While our society faces many challenges, those often least spoken about due to lack of clarity or 'taboo' nature pose the most significant long-term problems. With around half of all Americans experiencing a mental health disorder at some point in their lives ~\cite{Kessler2005}, it is clear that there is an ongoing mental health crisis. Psychotherapy, the practice of long-term conversational sessions with trained professionals, has been an potential remedy that has been shown to provide some respite for those suffering ~\cite{Lambert2002}. It is important to note that this paper sometimes uses the term counselling interchangeably with psychotherapy. While there are differences between the two terms, namely in the duration of the specific therapy and training required for the therapeutic intervention, our paper will not be focused on the topic for which the differences exist and hence will not make distinctions between them. 

It is important also to mention that just because mental health treatment such as psychotherapy exists, it does not mean all can receive access to it. In 2019, over 60\% of people under 18 in America with major depression did not receive mental health treatment \cite{Hellebuyck2019}. Unfortunately, the mental health crisis has been further exacerbated by the COVID-19 pandemic, which is beyond just creating anxiety and distress ~\cite{Nelson2020}. It has also led to systemic changes likely to have long-term impacts. An example is an increase in reclusive behaviour, shown through the popularisation of remote working environments, with projections of the number of Americans working remotely in 2025 at an increase of 87\%  compared to before the pandemic \cite{UpworkStudy}.

Arguments are being made that this increase in isolation, among other prevailing impacts of the pandemic, will have severe downstream effects on mental health in the following years. Many are looking to technology for a more effective and accessible solution than we have relied upon previously. 

\subsubsection{Technology In Mental Health}

Technology has prompted various approaches to address mental health illnesses, from Telehealth to Virtual Reality Immersion Therapy to Mobile Application Self Help. The utilisation of technology has led to reductions in the costs of therapy and an improvement in the ease of access. It has indirectly reduced the stigma associated with attending treatment ~\cite{Ralston2019}.

The latest introduction of Generative AI is one that many believe could revolutionise the Healthcare industry. It provides exciting promises of more accurate diagnoses, enhanced accessible global support, and the ability to study rare cases in greater detail. A component of Generative AI that could be the key to providing personalised support at scale for mental health is that of Large Language Models (LLMs). However, if LLMs are to be used in Mental Health, we must understand them and their risks exceedingly well.

\subsection{Matter Vs. Mode}

In this section, we emphasise the importance and difficulties that come with attempting to understand and quantify the subjective side of communication. We provide an overview of a subdomain of psychotherapy that we believe holds a lot of potential value. We will use this later on in our framework and metrics to help provide ways to measure nuanced conversation abilities.

\subsubsection{It's Not What You Said, It's How You Said It}

Due to the subtle and subjective nature of human communication, it's hard to describe the best way to communicate effectively to someone. Based on personal preferences and experiences, some argue conveying more emotion is better, while some would emphasise the importance of being direct. Some may articulate the need to ask many questions, while others say taking over the conversation is essential. We will refer to the strategies that enable subtle, context-appropriate, and intelligent communication as `nuanced' communication skills. We can establish this term to represent a combination of linguistic techniques, such as knowing when to ask questions or agree with someone and knowing effective communication, i.e., the study of emotional intelligence in communication.

When considering future applications of LLMs in mental health, as discussed above, it is clear that we need to have a good idea of the nuanced conversation abilities of these LLMs. This is due to the nature of the discussions these LLMs will be having with users regarding sensitive topics such as severe major depression, suicide, or substance abuse. A lack of adequate consideration of emotion, sentiment, or communication strategies may lead to tragic consequences that could be averted.

\subsubsection{Conversation Analysis}

%We previously mentioned how nuanced communication utilises linguistic strategies and affective knowledge. 
We leverage a sub-field of mental health and psychotherapy and conversation analysis better to understand linguistic techniques and strategies for better communication.

In psychotherapy, conversation analysis studies the communication between a therapist and a client. It helps us understand the aspects of communication that we intuitively use but might need help explaining concretely. In this field, we see this used to identify the best ways to connect with and help clients. The techniques of conversation analysis are leveraged to improve patient outcomes. An example is formulations, which provide a framework through which a therapist can best respond to a client's points \cite{Weiste2013}. A specific example of an aspect discussed in conversation analysis is lexical substitution. The therapist typically does this to heighten the emotion of a description provided by a client. It shows attentiveness and works to encourage the client to talk about their feelings and expand more than they initially would have \cite{Vehviläinen2008}.

We will draw from this field to identify specific means and techniques of communication used by psychotherapists in real life. We later leverage this knowledge to evaluate the nuanced conversation skills of LLMs, which will help provide a way to measure attributes of LLMs that could highly influence on client well-being.

\section{Literature Review} 

The topic of Large Language Models (LLMs) is currently incredibly popular. At the time of writing, a search for 'Large Language Models' on Google Scholar (a source of scholarly papers) generates over 6 million results. For some contexts, this results in more than the amount generated by words like 'Machine Learning, ' 'Artificial Intelligence, ' and 'Natural Language Processing. '

However, as we begin to peel back the layers and examine the existing research related to evaluating LLMs, the work concerning conversation abilities is limited. This is even more so the case once we narrow the research to that which relates to mental health.

This section will study the existing literature that has informed our work. We will segment this into multiple sections, draw key insights, and identify existing gaps. Due to the popularity of the field, it is possible that more related works have been published since writing. However, to the best of our knowledge, we have effectively tied in the key and notable works at the time of writing.

\subsection{LLM Conversation Evaluation} 

While lots of effort has been allocated to evaluate different types of LLMs for topics ranging from logical reasoning \cite{Liu2023} to trustworthiness \cite{Wang2023}, we explore relevant existing works that study conversation performance research.

We see papers in this domain tend to include partial discussions on this topic rather than focus an entire paper on it, such as in \cite{Bang2023}. While this paper provides an effective combination of human and automatic evaluation methods, the automated evaluation metrics are limited and are popular general ones used for LLMs, such as ROUGE \cite{Lin2004ROUGEAP}. Because of this, they lack the depth that applied conversation metrics would offer. The work also lacks a human baseline, which would provide an insightful perspective to compare performance. We see works that dedicate themselves to addressing this need by proposing complete automatic evaluation frameworks for this \cite{Lin2023}. However, the metrics used in this paper \cite{Lin2023} are grammar, context, and relevance. These are limited in terms of the number and depth provided by the analysis. It also tells us about the specific response from the LLM but less about the interaction between the LLM and the person. There also exist papers that push the whole field of analysing LLM dialogue forward by providing general large-scale conversation data \cite{Zheng2023}. These works provide high-quality data and explore the types of discussions that occur between LLMs and people. However, they don't study the interaction at a finer level to understand aspects such as how the LLM crafts the response and balances emotion.

We have attached a table from \cite{Chang2023} summarising the key existing benchmarks used for LLM evaluation. Of all the benchmarks mentioned, only one, EmotionBench \cite{Huang2023}, attempts to assess nuanced conversation skills at the level of interaction between LLM and person. In this case, it studies explicitly the affective nature of responses.

%\begin{figure}[h] 
%    \centering
%    \includegraphics[width=1\textwidth]{thesis/litreview/Sections/Benchmark Table.png} 
%    \caption{Current LLM Evaluation Benchmarks}
%\end{figure}

A limitation of this study is that it tends to focus on self-reporting for emotion analysis, which introduces self-reporting bias. A better approach would be to analyse and identify the emotion present within the text. Additionally, it focuses on assessing how LLMs may come up with and suggest appropriate emotions to given imagined situations. This differs from determining how they might naturally generate emotions in conversation. While this paper is a good start and provides some information regarding LLMs' nuanced affective conversation ability, it must be more comprehensive.

Based on the existing work we analysed, we have identified a gap regarding LLM's natural nuanced conversation abilities. While some related literature that discusses evaluation exists, the work needs more depth and applied metrics. Additionally, the existing work focuses more on question-answering performance than linguistic conversational abilities.

\subsection{Mental Health LLMs}

In this section, we identify and analyse notable papers combining mental health and LLMs. It is important to note that the number of relevant works is limited, which may be due to several complicated factors ranging from medical data privacy to the popularity of other domains being more appealing.

Arguably, the original paper that spurred interest in this domain proposed two of the first pre-trained models specialised for mental health use, MentalBERT and MentalRoBERTa \cite{Ji2021}. %The paper provides a look into a field that, at the time, had not been considered. 
However, an issue that is present in the paper is the use of unverified Reddit data to train the model. When evaluating the model, which is the topic we are most interested in, we see the paper focus on benchmarking the mental health detection abilities of the models. %A focus on evaluating mental health detection performance is a trend we continue to see across papers. The appeal of this is understandable. 
Having a chatbot that could have a conversation with an individual and then potentially diagnose them could be incredibly beneficial for early detection and treatment. 

While some relevant papers also provide attention to evaluating for demographic or racial bias, the focus tends to remain on detection performance \cite{Heinz2023}. This paper leverages a more popular and commonly used frontier model, GPT3, to conduct its analysis, making results more applicable than analyses done with MentalBERT or MentalRoBERTa. This is due to the widespread use and popularity of GPT3. However, a lack of depth is present by not comparing results across models. On top of this, by not providing any analysis of user interaction, detection performance is effectively done in isolation. The reality is that this separation from user interaction makes the results less meaningful, as any real applications that leverage the results will not be done in isolation from users. We see another relevant paper, \cite{Nguyen2022}, which focuses entirely on depression detection and investigates and attempts to improve the domain generalisability of models. The paper has promising results in identifying that grounding the model on a specific set of depression symptoms improves generalisability. Intuitively, we are also improving model interpretability by anchoring the model to a set of symptoms. However, this paper does use unverified Reddit data, which raises questions about how it will perform when applied in the real world. Due to the topic, it also focuses its evaluation solely on detection performance. Which, as aforementioned, needs more nuance and reveals little about user interaction. While the papers listed above tend to focus on a singular model and are focused on investigating one specific thing, we do see documents that are much larger in scale \cite{Xu2023}. This work evaluates a comprehensive amount of models, ranging from Alpaca to GPT4, across different categories, from zero-shot prompting to instructional fine-tuning. However, even this paper uses public data from social media sites to train the models. This opens the door to conversations about extrapolating these results when dealing with genuine data. Additionally, as mentioned in the paper and similarly to the documents listed above, the evaluation is focused solely on detection performance. 

When we diverge from the papers where the evaluation centres around mental health detection or classification \cite{Ji2021}, the literature becomes increasingly limited and relevant to our topic. We see papers put forward measures like word count per therapy session, sentiment analysis of the user, and emotion analysis of the user to measure LLM performance \cite{Mansoori2022}. While this paper provides a practical summary of several related topics, its analysis focuses on the user instead of the LLM. To emphasise that, all of the metrics mentioned are focused on analysing the content produced by the user rather than that of the LLM. When looking for papers that specifically focus on evaluating the characteristics of the responses made by the LLM, the work gets even more limited. Metrics such as the LLM's ability to interpret, explore, and demonstrate emotional reactions have been evaluated \cite{Sharma2020}. An issue with this paper is that the data used is not verified genuine mental health data. However, something they do incredibly well is leaning into the mental health domain and borrowing from the field of mental health counselling to use metrics like interpretations and explorations. %To the best of our knowledge this is one of, if not the only, paper in this domain which does this.

Several works establish a need for additional work in developing mental health LLMs for purposes other than standard diagnosis or record keeping. ``Chatbots for psychiatric diagnosis can not achieve satisfying performance by simply collecting symptoms like questionnaires. Instead, they should be equipped with various professional skills, such as emotional support, to complete the diagnosis task effectively'' \cite{Chen2023}. While this paper brings to light what a number of the previous ones are lacking in terms of what should be considered, it has places where it could be improved. Firstly, it focuses just on depression detection and doesn't include other conditions or topics. Secondly, while it discusses and evaluates several more interesting conversational characteristics like fluency, expertise, and engagement, these are reviewed by humans. This makes the evaluation less scalable and more open to bias. However, understandably, the nuance of the metrics lends to the difficulty in making them automatic. While we traditionally relied upon human evaluation to help judge hard-to-quantify nuanced qualities, we now see LLMs being leveraged for this task. For example, we have specifically seen papers leverage an LLM, RoBERTa, to help measure empathy \cite{Amjad2023}. While this is a significant step forward, this paper would benefit from tuning the model's temperature to make it increasingly deterministic among responses. It would also benefit from using frontier high-performance models for detection rather than just RoBERTa. Additionally, comparing performance across the chosen models would add an exciting facet of analysis.

Considering this grouping of literature, we can identify some things that are missing or could be improved in future work. Papers would benefit from an increased focus on evaluating the linguistic qualities of the responses provided by the LLM instead of assessing the LLM's ability to diagnose and classify the user. This is an avenue that has yet to be thoroughly studied. While difficult due to the nature of medical data, the quality of derived results would benefit highly from using verified mental health data. Most papers mentioned here use public open-source and unverified data from Reddit \cite{Sharma2020}, \cite{Amjad2023}. Additionally, it would be interesting to see a study across topics and different frontier LLMs to investigate how mental health topics impact LLM conversation ability. This is because some of the most relevant literature focuses on singular topics like depression \cite{Chen2023} and uses just one LLM in the analysis. Additionally, leveraging LLMs over human evaluations to measure nuanced metrics could provide a scalable and more objective alternative. Finally, leaning into the counselling world to derive more applied metrics, similarly to as started by \cite{Sharma2020}, could make generated results much more effective.

\subsection{Frameworks}

The following section will review existing papers that propose relevant frameworks to evaluate LLMs. Additionally, based on the papers, we will highlight areas where we believe improvement can occur.

The existing frameworks tend to focus on broad, high-impact topics. Another work puts forth a framework to assess the value of LLMs in Healthcare \cite{Reddy2023}. To complement the standard LLM evaluation metrics, such as BLEU \cite{Papineni2001} or ROUGE \cite{Lin2004ROUGEAP}, this paper puts forth several conceptual ones, such as Accountability or Trustworthiness. While this framework is interesting and provides a good lens through which to approach the adoption of LLMs in Healthcare, the conceptual metrics make it harder for users to apply them. Even a high-level explanation of a potential technical implementation process could make the framework more helpful to users. Another relevant paper provides a framework for implementing AI systems in healthcare \cite{Reddy2021}. This paper has put forward a framework that can be applied to other real-world systems. This paper mentions something of high importance: "One major limitation of existing reporting or evaluation frameworks is their narrow focus". While the specific framework isn't directly relevant to our work in mental health, this approach of having a framework that can translate to other relevant domains has high potential and will be something we utilise ourselves.

A highly relevant framework, which we mention in the previous section of this literature review, is one used for measuring empathy \cite{Amjad2023}. While this framework is great in theory and is based on solid theoretical grounding in justified emotion scales, some weaknesses could be improved. The paper uses a public Reddit dataset to train its model. This lack of verified data introduces the potential for bias. Additionally, they compared LLM results to those produced by humans to evaluate performance. The issue with this is that empathy is subjective and will vary from user to user; this may likely lead to inconsistent results. 

A big learning is the value of providing a conceptual framework that provides technical implementation information. This makes it more easily implementable. Additionally, we can adopt the idea of making the framework translatable to similar applications outside the mental health domain. This will lead to more people being able to make use of it. Finally, we identified that we can distinguish our framework through a baseline developed from high-quality data rather than subjective human judgments that vary. \cite{Amjad2023}.

\subsection{Summary of The Identified Gaps In The Literature}

Based on the existing literature, here are the identified gaps:

\begin{itemize}
  \item In the mental health domain, insufficient work has analysed LLMs at the point of interaction with humans.
  \item There is a lack of existing LLM evaluation metrics that leverage mental health conversation analysis literature.
  \item Not enough work has been done in assessing how LLM performance varies across mental health topics.
  More work must be done to compare LLMs' nuanced conversation abilities.
  There needs to be a relevant framework with implementation details that can be translated to relevant adjacent domains.
  \item Existing frameworks and mental health model evaluations do not use verified data to train and evaluate models.
  
\end{itemize}

\section{Methodology} \label{chap:methodology}

%\subsection{Introduction}

In this section, we present our methodology and link it to our research objectives. %, and how researchers can make use of it.
Firstly, we propose a succinct framework which can be used to effectively evaluate the nuanced conversation abilities of LLMs. We further cover data pre-processing, result generation, and analysis. This framework will be developed in a transferable style that is not specific to the domain of mental health, which will help us answer the following research objective: 

%\textbf{RO1}. To propose a novel evaluation framework which works to analyse the nuanced conversation abilities of Large Language Models. 

\textbf{RO1}: Inside the framework, we will develop and propose a number of metrics which range from analysing the affective abilities of LLMs to studying the strategies in which they answer questions. While some of the metrics will leverage mental health conversation analysis literature, they will be transferable to any domain where understanding nuanced conversation abilities is important. This will help us answer the research objective:

%\textbf{RO2}. To develop a number of quantitative metrics which vary from measuring affective content to emulating conversation strategies from counseling literature. 

\textbf{RO2}: The goal of our method is to clarify the selection of our course of action and to enable them to apply our framework to whichever LLM or context may suit any environment. 

%Here is a brief description of the key sections of the framework:
%\begin{itemize}

%    \item \textbf{Data Pre-Processing:} In this section we discuss how other researchers in the field may prepare data to use in the following sections of the framework.

%    \item \textbf{Metrics:} In this section we spend some time discussing in detail the metrics we selected to evaluate the nuanced conversational abilities of the LLMs. We provide a high level overview, a justified explanation, and a more technical implementation guide for each of them.
    
%    \item \textbf{Running The LLMs:} Here we provide information regarding the way in which we recommend researchers use their selected data and run their LLMs.

%    \item \textbf{Analysing Results:} In this section we provide recommendations on ways to go about analysing results produced from the metrics.
%\end{itemize}

\begin{figure}[ht] 
    \centering
    \includegraphics[width=1\textwidth]
    {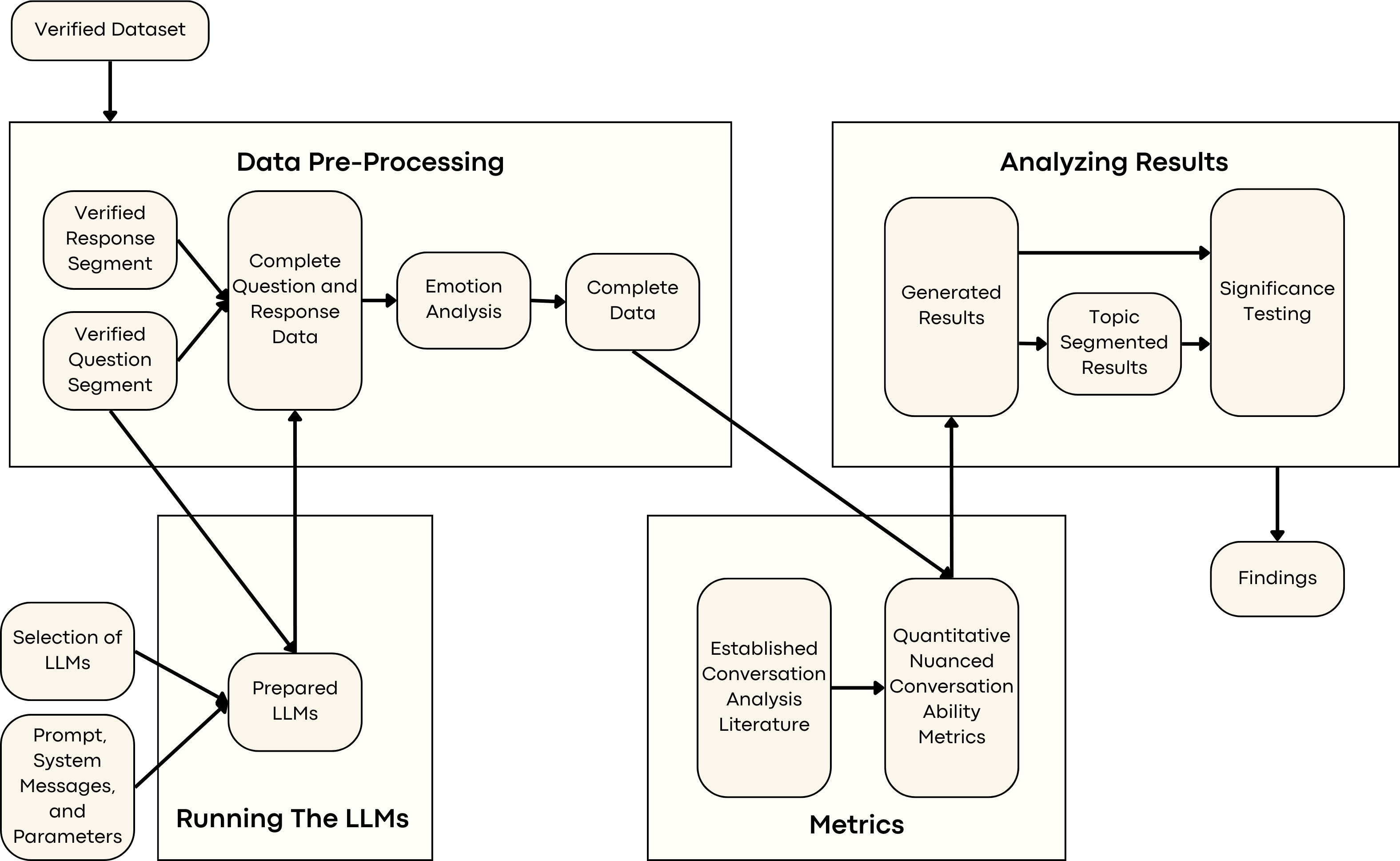} 
    \caption{Framework Overview. \label{framework}}
\end{figure}

Figure \ref{framework} presents a high-level overview of our framework. We provide a detailed discussion on various components of our proposed framework below. 

\subsection{Data Pre-Processing}

It is highly recommended to identify a high-quality verified dataset; however, in some domains, such as healthcare, this may be hard due to privacy or regulatory reasons. %However, having verified genuine conversations to analyse will lead to results that hold more veracity. 
Depending on the context, we recommend the data be collected from verified professionals or experts in the relevant domain, e.g., doctors in healthcare or investment professionals in finance. Additionally, we ensure that the dataset format consists of a collection of questions and answers. There are datasets where each question has multiple answers; hence, we recommend identifying which answer is the most popular or of the highest quality. This may be done by filtering by metrics (e.g., `likes' or `upvotes') 

Following this, we suggest splitting up the verified questions and the answers. This dataset of answers will form our `gold standard' (GS) baseline, representing what an ideal response should look like. Because of this, having access to verified data, consisting of responses from professionals, is critical. We will then feed the set of questions into selected LLMs to collect their responses to each question. In addition, we advocate using an emotion identification process, specifically using an LLM trained on emotion data, to classify the emotions conveyed in each of the questions, GS baseline responses, and LLM responses. This process will enable us to have the following collected and processed data:
\begin{itemize}
    \item A verified question
    \item The emotions present in the verified question
    \item A verified gold standard response
    \item The emotions present in the verified gold standard response
    \item A response from each of the LLMs used
    \item The emotions present in each of the LLM responses
\end{itemize}

We additionally recommend using standard data processing and cleaning procedures for text.

\subsection{Metrics}
We now describe the metrics we have identified and developed to provide insights regarding the nuanced conversation abilities. The metrics will be presented in the following format:
\begin{itemize}
    \item High-level explanation of the metric
    \item Justification of the metric
    \item Information regarding previous metric implementation and limitations
    \item Pseudocode to help guide implementation from other researchers
\end{itemize}

It is important to note that these metrics are transferable to other adjacent domains where nuanced conversation is essential. We will be leveraging mental health counselling literature to develop the following metrics. This is because conversation analysis, a subfield of counseling, provides systems for understanding conversation strategies at a deeper level. The value of these strategies will help us provide quantitative ways to measure some of the subjective aspects of conversation.

%We will now provide the framework to apply to the data we have collected and pre-processed. The framework's goal is to provide a series of metrics which provide clarity regarding, and enable us to measure some of, the nuanced nature of communication. 

%While we be applying this framework to this context of mental health, the framework can be applied to other LLM interactions where the way in which information is provided is highly important. A quick example to highlight this would be a future application of using multi modal LLMs to respond to emergency calls from people. (More on future extensions later on though).

%We were able to create a number of the metrics by leveraging insight provided from the field of conversation analysis, as its deep study into effective communication proved hugely informative.

\subsubsection{Emotion Consistency}

%\textbf{High Level Explanation}: 
This metric works to assess the similarity and overlap between the emotions present in verified GS (`gold standard') responses and those present in the LLM responses. It enables us to determine how closely an LLM is able to generate the same emotions present in GS responses, when responding to verified questions.

\textbf{Metric Justification}: Leveraging emotion in conversation is a powerful tool yet it is hard to concretely describe. The importance of effective emotion use in the field of mental health is made clear in Relational Integrative Psychotherapy \cite{Finlay2015}, ``Empathy and Attunement underpins therapy''. However, the complex and nuanced nature of these topics is also established in the cited work by their descriptions as ``elusive processes and hard to describe''. The importance of understanding and reflecting feelings is further discussed in counseling handbooks \cite{Nelson-Jones2005}, where the authors specifically say that ``Reflecting clients' feelings at the start of initial sessions shows that one is tuned into them as persons''. Based on the cited literature, understanding and using emotion properly is a key aspect of nuanced conversation. %It is due to this that we will develop a metric for it.

\textbf{Previous Metric Use and Limitations}: Due to the importance of understanding and demonstrating emotions in certain safety critical contexts, such as mental health or first-responder calls, testing the abilities of LLMs is not a novel concept. Specifically in the field of mental health, e.g., \cite{Sharma2020}, \cite{Amjad2023} attempt to do this with empathy. However, while these papers used general labeled data to train models on, we are advocating the use of verified and domain-specific datasets which separately have emotion identification processes run on them. This makes the results more applied to the specific domain. Additionally, rather than look at empathy itself which is more subjective and hard to describe, we look at the emotions identified. We also take into account the ranking and weight of the emotions present in each response. This is a novel adaption which provides more depth to the analysis. A limitation of our metric is that we are leveraging emotions identified by an LLM for our analysis. This means that the performance of our metric is dependent on the performance of specific LLM, which is important to consider and could be improved through the use of more than one LLM, or even including human evaluations, to cross check emotion identification.

\textbf{Metric Implementation}: In order to implement this metric, we made use of the weighted list of emotions present in the GS response, as well as those in each of the LLM responses. We also used, and recommend the use of, the Ranked Based Overlap (RBO) function \cite{Joshi2021} to derive similarity between the ranked lists of emotions. The Algorithm \ref{algo1} shows a simplified pseudocode implementation of this metric. 

\begin{algorithm}[ht]
\caption{Emotion Consistency}
\begin{algorithmic}[1]
\Require \textit{GS Emotions, LLM Emotions}
\Ensure \textit{Emotion Consistency Score}
\Procedure{Emotion Consistency}{GS Emotions, LLM Emotions}
    \State $Emotion Consistency Score \gets \text{RBO}(GS Emotions, LLM Emotions)$
    \State \Return $Emotion Consistency Score$
\EndProcedure
\end{algorithmic} \label{algo1}
\end{algorithm}

\subsubsection{Relative Directional Change In Sentiment}

%\textbf{Metric Explanation} 

This metric measures the relative directional difference in sentiment between questions and responses. It helps us identify how LLMs produce specific sentiments in response to the exhibited sentiment of questions. We will run this metric with the GS baseline response to provide a verified human baseline (i.e. to understand what appropriate changes in sentiment look like) and then compare the metric results for LLMs to it.

\textbf{Metric Justification}: Responding to a message appropriately requires an adequate understanding of modulating sentiments based on what is provided. If someone were to say something that indicated they were `Sad', we would respond differently than if they were `Happy'. The issue is this is incredibly subjective and depends on the person, the existing relationship, and the context. Due to this grey zone on what is needed, we can leverage the GS response sentiments as a baseline. Additionally, analysing sentiment provides a more holistic and continuous view of the situation than analysing emotions alone; this is due to the discrete nature of emotions. `Joy' and `Happiness' are considered two separate emotions in the same way that `Joy' and `Anger', despite the first pair being drastically more similar than the second; hence, sentiment provides more nuance, which we can learn from.

\textbf{Previous Metric Use and Limitations}: In the field of LLMs, sentiment analysis is an evaluation metric considered. However, the existing literature focuses predominantly on evaluating the ability of LLMs to conduct sentiment analysis. This is a famous line of existing research, as seen in Chang's survey \cite{Chang2023}. This stands in contrast to our intended evaluation, which is to do sentiment analysis on the text produced by the LLM. A limitation worth considering is the subjective nature and the context dependence. People respond to situations in various ways; some are sarcastic, while others are overly empathetic. This means the results identified are highly entrenched with the dataset and specific domain applied. It is also linked to the demographic of the participants in the dataset.

\textbf{Metric Implementation}: Algorithm \ref{algo2} presents insight into the implementation of this metric. We suggest using the TextBlob package \cite{loria2018textblob} to utilise a sentiment analysis function that returns the polarity of a piece of text. However, researchers may use another sentiment analysis substitute that better suits their needs. Using the percentage difference in Algorithm \ref{algo2} enables us to understand the relative change in sentiment between the questions and responses. However, the total value can be huge if the question has a sentiment near 0. This is one of the reasons we then recommend normalising the values to the range of 0 to 1.

\begin{algorithm} [ht]
\caption{Relative Directional Sentiment Change}
\begin{algorithmic}[1]
\Require \textit{Text}
\Ensure \textit{Sentiment Polarity of Text}
\Procedure{Analyse Sentiment}{Text}
    \State $Sentiment of Text \gets \text{Texblob Sentiment}(Text)$
    \State $Sentiment Polarity of Text \gets Sentiment of Text.Polarity$
    \State \Return $Sentiment Polarity of Text$
\EndProcedure

\Require \textit{Question, Response}
\Ensure \textit{Normalised Change In Sentiment}
\Procedure{Sentiment Change}{Question, Response}
    \State $QuestionPolarity \gets \text{AnalyseSentiment}(Question)$
    \State $ResponsePolarity \gets \text{AnalyseSentiment}(Response)$
    \State $Change \gets \text{PercentageDifference}(QuestionPolarity, AnswerPolarity)$
    \State $Normalised Change In Sentiment \gets \text{SigmoidFunction}(Change)$
    \State \Return $Normalised Change In Sentiment$
\EndProcedure
\end{algorithmic} \label{algo2}
\end{algorithm} 

\subsubsection{Intra Response Sentiment Change}

%\textbf{Metric Explanation} 

The previous metric measures changes in sentiment between questions and responses; this metric will be used to identify the changes in sentiment throughout just a response. The common practice of using a single sentiment value to summarise an entire response must include more helpful information. This will help provide more insights into it. It will enable us to compare sentiment change throughout a GS response to sentiment change throughout an LLM response. To provide some quick intuition for what to expect, this metric will have a high value if a response becomes increasingly positive throughout it.

\textbf{Metric Justification}: Leveraging existing conversation analysis literature, we can gain some understanding of how effective conversational responses are constructed. In conversation, counsellors are recommended to do things such as reflect on patients' feelings. This is discussed in the practical counselling handbook \cite{Nelson-Jones2005}, e.g., ``[clients] may feel better understood by trainees who reflect their main feeling at the front of their response than if they reflect information first''. We have also seen the need for multiple components to exist in a response expressed more explicitly, ``Next turns are understood by co-participants to display their speaker's understanding of the just prior turn and to embody an action responsive to the just-prior turn so understood.'' \cite{Schegloff2007}. In summary, good responses require several different actions in different stages. Because of the variety of things expected in a high-quality response, understanding changes in sentiment over the response could be highly beneficial. This justifies our interest in measuring how LLMs compare to a verified baseline when doing this.

\textbf{Previous Metric Use and Limitations}: The authors \cite{Syzdek2020} leveraged sentiment analysis to derive insights from counselling sessions. The paper provides high-quality insight at a more macro scale (i.e. analysing sentiment over conversations and counselling sessions) rather than at the micro-scale we want to study (i.e. in responses). They mention they average the sentiment across the response, ``the sentiment of a particular turn was assessed by obtaining the mean of the sentiment scores of individual sentences within that turn''. We will be able to provide insights at a more granular level. A limitation of this metric is that the longer the responses, the more information we get from the metric. %This is something to keep in mind. 
Because of this, the metric may vary depending on context, e.g., in a first responder call where responses may need to be short, this metric may be less useful. Whereas, in an educational setting where responses may be more detailed with multiple sentences, this metric may provide lots of helpful information. 

Due to the popularity of conducting sentiment analysis, there is potential for more similar work to exist. However, to the best of our knowledge, evaluating this at an intra-response level within the context of LLMs is novel.

\textbf{Metric Implementation}: This metric was implemented by splitting responses into sentences and then calculating the sentiment of each sentence. We then took those sentiment values, added time values to correspond to their locations, and then calculated the linear regression of the appropriate line. We then took the slope of this line to represent the change in sentiment over the message. We also normalised the value to ensure consistency across a range of metrics. 

Algorithm \ref{algo3} shows this metric implementation. It is also worth noting that we utilise the \textit{analyse sentiment} function from the previous metric for this metric.

\begin{algorithm} [ht]
\caption{Intra Response Sentiment Checking}
\begin{algorithmic}[1]
\Require \textit{Response}
\Ensure \textit{List Of Sentiment Over Response}
\Procedure{IntraSent}{Response}
    \State Declare \texttt{SentimentValues} as an empty list
    \State $ListOfSentences \gets \text{SplitTextForEachSentence}(Response)$
    \For{each $\textit{Sentence}$ in $\textit{ListOfSentences}$}
        \State $SentimentValues.Append(\text{AnalyseSentiment}(Sentence))$   
    \EndFor
    \State \Return $SentimentValues$
\EndProcedure
\Require \textit{List Of Sentiment Over Response}
\Ensure \textit{Normalised Change In Sentiment Across Response}
\Procedure{LinearRegressionSent}{List Of Sentiment Over Response}
    \State $LinearRegressionModel \gets \text{LinearRegression}(ListOfSentimentOverTime)$
    \State $SlopeOfModel \gets \text{Coefficient}(LinearRegressionModel)$
    \State $NormalizedSlopeOfModel \gets \text{Sigmoid Function}(SlopeOfModel)$
    \State \Return $NormalizedSlopeOfModel$
\EndProcedure
\Require \textit{Response}
\Ensure \textit{Normalised Change In Sentiment Across Response}
\Procedure{IntraResponseSentiment}{Response}
    \State $ListOfSentimentOverResponse \gets \text{IntraSent}(Response)$
    \State $NormalizedSlope \gets \text{LinearRegressionSent}(ListOfSentimentOverResponse)$
    \State \Return $NormalizedSlope$
\EndProcedure
\end{algorithmic} \label{algo3}
\end{algorithm} 

\subsubsection{Simplicity}
%\textbf{Metric Explanation} 

In this section, we measure the response's simplicity and effective readability. 

\textbf{Metric Justification}: The practical counselling guide takes a straightforward stance on the importance of simple and readable language when responding to questions, ``Use simple and clear language. Avoid unnecessary words and qualifications.'' \cite{Nelson-Jones2005}. Due to the established importance of simplicity, we can assure that there is value in analysing how the simplicity of LLM responses compares to that of verified GS responses.

\textbf{Previous Metric Use and Limitations}: In existing LLM research, simplicity has often been measured when dealing with cases like text simplification (summarising tasks) and the output of a translation (translating from one language to another). In cases like this, it has been measured through human evaluators and metrics such as BLEU \cite{Papineni2001}. We believe that analysing simplicity in more general responses could prove insightful. We also believe that the reactions can be evaluated with simple metrics that are not LLM-specific. A limitation is that it is known for simplicity and readability scores to suffer over shorter texts. Given the fact the text we are analysing is relatively straightforward, this is a severe limitation to the results we will achieve. 

\textbf{Metric Implementation}: %We encourage researchers using our framework and metrics to use whichever simplicity metric is most appropriate for their context. If no specific metric is needed, we recommend using the Flesch-Kincaid reading ease metric, which can be accessed through the Python Readability package \cite{ReadabilitySoftware}. 
The American military used the Python Readability package \cite{ReadabilitySoftware} metric to test the readability of their manuals and guides. It returns a score from 0 to 100; the lower the score, the less simple the text and the more difficult it is to read. The issue with this simplicity metric is that it requires text to be over 100 words. Due to this, we filter out text under 100 words for this metric. The Algorithm \ref{algo4} shows the pseudocode of a sample implementation for this metric. 

\begin{algorithm}[ht]
\caption{Simplicity Element Checking}
\begin{algorithmic}[1]
\Require \textit{Filtered Text}
\Ensure \textit{Flesh-Kincaid Ease Of Reading Score}
\Procedure{KincaidScore}{Filtered Text}
    \State \textit{Results} $\leftarrow$ Readability(\textit{Text}, 'English')
    \State \textit{Flesh-Kincaid Ease of Reading Score} $\leftarrow$ \textit{Results}['Readability Grades']['Kincaid']
    \State \Return \textit{Flesh-Kincaid Ease Of Reading Score}
\EndProcedure
\Require \textit{Response}
\Ensure \textit{Flesh-Kincaid Ease Of Reading Score}
\Procedure{Checking Simplicity}{Response}
    \State $NumOfElements \gets \text{LengthSplitElements}(Response)$
    \If{$ {\textit{NumOfElements}} <= 100$}
        \Return $Nothing$
    \EndIf

    \State \textit{Flesh-Kincaid Ease Of Reading Score} $\leftarrow$ \Call{KincaidScore}{Response}
    \State \Return \textit{Flesh-Kincaid Ease Of Reading Score}
\EndProcedure
\end{algorithmic}\label{algo4}
\end{algorithm} 

\subsubsection{Recycling Elements}

%\textbf{Metric Explanation} 

This metric measures the number of words a response recycles and reuses from a question. 

\textbf{Metric Justification}: In the field of conversation analysis, the concept of a highlighting formulation exists to instruct therapists on how to reuse elements provided in client dialogue. One of the leading papers on this topic describes: ``In a highlighting formulation, the therapist selects a part of the client's prior turn, recycling some of its key descriptive elements ... By formulating these key descriptions, the therapist shows that (s)he has listened to and understood the client’s description.'' \cite{Weiste2013}. Demonstrating and expressing understanding simply through the reuse of key descriptive elements is incredibly powerful. Hence, it is important to propose a way to evaluate the amount of recycling done by the GS baseline and analyse how our chosen LLMs compare to it.

\textbf{Previous Metric Use and Limitations}: Retroactively analysing sensitive conversations, like counselling ones, and applying principles of conversation analysis is presented in \cite{Madill2001}, is not a widely used practice. Specifically, to the best of our knowledge, within this context of evaluating LLM conversational abilities, we have not seen this implemented before. Developing a metric to analyse this formulation from the study of psychotherapy conversation analysis will be highly beneficial. One of the limitations of the metric is having to filter our irrelevant, easily recyclable `fluff' words that do not contribute meaningfully. These words may provide structure to the sentence or make it grammatically correct, yet do not provide any critical descriptive or explanatory information. These need to be removed from consideration since responses may easily recycle words like this due to language constraints. We recommend dealing with this by creating a list of `fluff' words and attempting to filter them out. The list of words will depend on the context in which the framework is being applied and how strict the researcher wants to be.

\textbf{Metric Implementation}: We have included a sample list of `fluff' words that were generated using ChatGPT3.5 Turbo. %Researchers are welcome to use this list or use their own.

fluff = set(["i", "me", "my", "myself", "we", "our", "ours", "ourselves", "you", "your", "yours", "yourself", "yourselves",
                  "he", "him", "his", "himself", "she", "her", "hers", "herself", "it", "its", "itself", "they", "them", "their",
                  "theirs", "themselves", "what", "which", "who", "whom", "this", "that", "these", "those", "am", "is", "are", "was",
                  "were", "be", "been", "being", "have", "has", "had", "having", "do", "does", "did", "doing", "a", "an", "the", "and",
                  "but", "if", "or", "because", "as", "until", "while", "of", "at", "by", "for", "with", "about", "against", "between",
                  "into", "through", "during", "before", "after", "above", "below", "to", "from", "up", "down", "in", "out", "on", "off",
                  "over", "under", "again", "further", "then", "once"])

Algorithm \ref{algo5} examines a sample pseudocode that can be used to implement the metric.

\begin{algorithm}[ht]
\caption{Recycling Element Checking}
\begin{algorithmic}[1]
\Require \textit{Question}, \textit{Response}
\Ensure Recycling Score
\Procedure{CheckingWordRecycling}{Question, Response}
    \State $UniqueQuestionElements \gets \text{SetOfLowercaseElements}(Question)$
    \State $UniqueNonFluffResponseElements \gets \text{LowercaseElementsNotInFluff}(Response)$
    \If{$\Call{Length}      {\textit{UniqueNonFluffResponseElements}} = 0$}
        \Return $0$
    \EndIf
    \State $\textit{RecyclingCount} \gets 0$
    \For{each $\textit{Element}$ in $\textit{UniqueNonFluffResponseElements}$}
        \If{$\textit{Element}$ is in $\textit{UniqueQuestionElements}$ and not in $\textit{Fluff}$}
            \State $\textit{RecyclingCount} \gets \textit{RecyclingCount} + 1$
        \EndIf
    \EndFor
    \State $Recycling Score \gets $\textit{RecyclingCount}$ \div \textit{UniqueNonFluffResponseElements}$
    \State \Return $Recycling Score$
\EndProcedure
\end{algorithmic}\label{algo5}
\end{algorithm} 

%To provide some more information regarding pseudocode, 
Note that we only calculated the set of elements (unique elements) from the question and not the set from the response. This is because if the responses reuse a crucial element from the question more than once, that should also be factored in. Note that it may choose to implement this differently based on the context of the investigation.

\subsubsection{Agreeability}
%\textbf{Metric Explanation} 

This metric investigates how much a response demonstrates agreement with the question. It is more nuanced than it might initially appear because more deals are not always better, contrary to what you may intuitively think. For example, sometimes, in the field of mental health, a therapist needs to push back and argue against what a client is asking. Here, we present examples highlighting the difference between Agreeability and Active Listening.% (the following metric). 

\textit{High Agreeability + Low Active Listening}: 
\begin{itemize}
    \item Message: ``I want to go buy a bike because I am sad''
    \item Response: ``Good idea!''
\end{itemize}

\textit{Low Agreeability + High Active Listening}:
\begin{itemize}
    \item Message: ``I want to go buy a bike because I am sad''
    \item Response: ``It sounds like you want to distract yourself by spending money, that is a bad idea''
\end{itemize}

\textbf{Metric Justification}: We will again leverage the mental health field to examine how understanding agreement factors into effective responses. Finding the right balance between agreement and disagreement is something many therapists need to do to help patients. The need for disagreement may seem surprising, but research has been released that disputes do not necessarily lead to worse outcomes \cite{Holmqvist2016}. An example mentioned in this referenced study of how this manifests is agreement over the existence, or nature, of problems faced by a patient. Evaluating the extent to which LLMs can produce appropriate levels of agreement, as determined by proximity to the GS baseline, clearly has value. 

\textbf{Previous Metric Use and Limitations}: The value of understanding Agreeability has been highlighted in the past. However, we have seen it measured more as a personality trait \cite{Hirsh2012}. This contrasts how we are trying to measure it, as demonstrated in conversation. The issue is the nuance that Agreeability takes. Words of affirmation can quickly have a reversed meaning depending on context and if they come across as sarcastic. This nuance makes it a prime candidate for classification by an LLM trained on a large amount of text. It is important to note that the limitation of this approach is any potential bias in the training data could produce worse results. Hence, we recommend cautiously selecting the LLM for this task. Another limitation is that results will vary depending on the exact wording of the prompt and system message. This may disadvantage people who are writing the prompt in a language they are not fluent. %This is something to keep in mind for researchers.

\textbf{Metric Implementation}: We will now provide some metric implementation information for researchers to use in their studies. As mentioned earlier, we recommend tailoring the method to the context. We also suggest the use of ChatGPT3.5 Turbo for this metric. %However, at the time of reading there may be a more appropriate model, hence, we recommend researchers make use of it. 
Following, we provide a sample system message and prompt for the model that researchers can use. We also present sample pseudocode in Algorithm \ref{algo6}.

System Message = ``Respond to the prompt exactly and provide strictly a number. With no clarification.''

Prompt =  ``From 0, which indicates strong disagreement, to 10, which indicates strong agreement, give a rating for the level of agreement provided by the second text for the first text. I want to define agreement as support and agreement presented in the message. This will be witnessed through linguistic choices. You may base the rating off the general message present in each and key phrases and words indicating agreement or disagreement.''

\begin{algorithm}[ht]
\caption{Agreeability Check Using LLM}
\begin{algorithmic}[1]
\Require \textit{Question}, \textit{Response}, \textit{Prompt}, \textit{System Message} 
\Ensure Agreeability Score
\Procedure{RunLLM}{Prompt, Question, Response, System Message}
    \State $ChatPrompt \gets Systemperature Message + Prompt + Question + Response$
    \State $GPTResponse \gets \text{Send To GPT3.5}(ChatPrompt)$
    \State $Agreeability Score \gets \text{ExtractScore}(GPTResponse)$
    \State \Return \textit{Agreeability Score}
\EndProcedure
\end{algorithmic}\label{algo6}
\end{algorithm}

\subsubsection{Active Listening}
%\textbf{Metric Explanation} 

This metric is focused on measuring active listening. For our paper, we define this as the content provided by a question, which is then sent back in a response. To clarify, this is slightly different than sending back recycled words because the content has to do more with substance than phrasing. Additionally, this does not consider agreement with the content but just whether it is present in the message sent back. To explain this better, we present examples highlighting the difference between Active Listening and Agreeability (i.e., the previous metric).

\textit{High Agreeability + Low Active Listening}: 
\begin{itemize}
    \item Message: ``I am so angry, my dad is making me do my homework''
    \item Response: ``It makes sense you feel that way''
\end{itemize}

\textit{Low Agreeability + High Active Listening}:
\begin{itemize}
    \item Message: ``I am so angry, my dad is making me do my homework''
    \item Response: ``I think you are wrong to feel angry about your dad making you do your homework, he is doing it because he cares about you''
\end{itemize}

\textbf{Metric Justification}: We can lean on mental health to understand the importance of evaluating active listening in responses. A practical counselling guide emphasises the importance and potential value of reflecting reasoning and not just specific words: ``A useful variation in active listening is to reflect both feelings and the reasons for them.'' \cite{Nelson-Jones2005}. This method of selecting what the patient has said and reshaping it in the response is a type of formulation. The impacts of employing this strategy during mental health-related conversations range from preparing topics mentioned by patients for further analysis to helping guide overall interactions with patients \cite{Vehviläinen2008}. Due to the importance of demonstrating active listening in conversation, we can recognise the value that evaluating it in the LLM responses will provide. This will especially be the case when we compare the active listening in the LLM responses to that of the GS baseline.

\textbf{Previous Metric Use and Limitations}: We noticed that active listening has become a topic of interest for chatbot performance in adjacent fields. However, given the subjective nature of active listening, it has been a metric where evaluation has suffered. Few papers conduct an in-depth study into building interview chatbots with functional listening skills \cite{Xiao2020}. However, this paper uses metrics such as the participant's reviews and continued interactions with the chatbots (i.e., the time spent and their response metadata) to measure the active listening of the chatbot. The issue with this method is the subjective nature of the reviews. Measuring engagement from users may be more telling of the participants' nature than the chatbot's active listening abilities. To measure the quality of the responses (and to sidestep the subjectivity issues from users), we recommend using an LLM, which has been trained on large amounts of text for active listening evaluations. As mentioned in the previous metric, using LLMs for evaluation introduces a set of limitations, including potential bias from training data to the output. Consequently, we recommend using this metric while selecting the LLM they would use with appropriate consideration.

\textbf{Metric Implementation}: We recommend carefully considering the prompts and system message selection to implement this metric. While we use GPT3.5, researchers should exercise caution when selecting the LLM they use for the implementation. The model's training data and several parameters can impact the quality of the results. Here is an example system message and prompt we recommend:

System Message = ``Respond to the prompt exactly and provide strictly a number. With no clarification.'' 

Prompt =  ``From 0 which indicates weak active listening, to 10, which indicates strong active listening, give a rating for the level of active listening between the following two pieces of text. Active listening involves how the second text captures the central message, topic, and reasoning discussed within the first text. Say for example, the first text says I suffer from depression, abuse, and divorce, and the second text says it can be challenging to suffer from multiple issues at the same time. This would constitute high active listening because while the words of the first text aren't repeated in the second, the central message is repeated. If in response to the first text, the second text just said I agree, that would be a low active listening score since it doesn't capture the content or reasoning of the first text. What matters is just whether the central messaging, topic, and reasoning is discussed in the second text.'' 

Refer to Algorithm \ref{algo7} for some pseudocode regarding a potential implementation strategy.

\begin{algorithm}[ht]
\caption{Active Listening Check using LLM}
\begin{algorithmic}[1]
\Require \textit{Question}, \textit{Response}, \textit{Prompt}, \textit{System Message}
\Ensure Active Listening Score
\Procedure{Run LLM}{Prompt, Question, Response, System Message}
    \State $ChatPrompt \gets System Message + Prompt + Question + Response$
    \State $GPTResponse \gets \text{SendToGPT3.5}(ChatPrompt)$
    \State $Active Listening Score \gets \text{ExtractScore}(GPTResponse)$
    \State \Return \textit{Active Listening Score}
\EndProcedure
\end{algorithmic}\label{algo7}
\end{algorithm}

\subsection{Running LLMs}

In this section, we briefly discuss how we recommend users of the framework apply the metrics to their chosen LLMs, with their selected data. Firstly, it is important to note that this process varies depending on the LLM model selected as well as %your chosen way of 
accessing it. We suggest spending enough time thinking about this in conjunction with their dataset size as well as the potential cost of running the models. Some closed models can be accessed easily with APIs, others can be downloaded and used locally, and some can be accessed on third-party sites.

The simplest way to generate the LLM responses is to loop through the dataset pass each question through to the LLM, and then collect the response. This was briefly discussed in Section 4.1. Following this, we recommend iterating through the dataset and running each metric on all of the LLM responses, and the GS-verified responses, for each row. For some of the metrics, we will need to also process the questions. Following this, we suggest averaging out the performance of each response for each metric. %It may be the case that some researchers should refrain from doing this in order to conduct different types of analysis. 
We note that some datasets provide additional detail that can then be used to assess performance across LLMs. If this is the case for a researcher's specific data, we suggest they maintain the value that indicates this through the processing, as a result, the data can be subset by the chosen features later on.

\subsection{Analysing LLM Results}

This section details how we suggest the results be analysed and evaluated. It is important to note that the depth of the analysis may vary depending on context and future planned applications of the results. Firstly, we make an important note regarding structural differences between metrics that impact how we can analyse them. We classify our metrics into two categories:

\begin{itemize}
    \item \textbf{Category 1}: Inclusive of the baseline. This includes metric 1, i.e., Emotional Consistency. Within the calculation of the metric, we make use of the GS baseline. Because of this, the metric itself already represents how close the LLM is to the baseline. Intuitively, the closer this metric is to the number 1, the better; the closer it is to 0, the worse.
    \item \textbf{Category 2}: Exclusive of the baseline. This includes the rest of our metrics. They do not include the baseline within their calculation, meaning the result produced still needs to be compared to the baseline to provide useful information.
\end{itemize}

In the following step, we recommend adequate data preparation before the analysis. It is important to note that due to the variation in metrics ranges, we suggest normalising them to the same range, i.e., 0 to 1. Other metrics, such as Agreeability and Active Listening, will need to be normalised by ten and depending on how Simplicity was calculated, it may need to be divided as well.

%When it comes to comparing how similar each LLM is to the GS across all the results from the Category 2 metrics, researchers are faced with another decision. 
Two approaches can be taken. While Euclidean Distance represents the straightforward exact distance across all the metrics, we recommend using the Pearson Correlation. This way, the directional relationship between the GS baseline and each LLM can be captured and compared. %However, this is a decision up to the researcher's discretion applying it and the context it is being used to. 
In addition, it is essential to ensure the identified correlations between some of the models and the baseline are statistically significant; for this, we recommend using P values. This will help simplify understanding of how likely a correlation between a specific LLM and the GS baseline is due to chance.
%The following advice is for researchers that have an additional independent variable that they wish to analyse LLM performance across. 
To study how specific LLMs perform when the independent variable is set differently, we suggest using Steiger's Z Test or a similar metric judged appropriately by the researcher. This will help identify whether superior performance from an LLM is due to chance when the independent variable is set to a specific condition instead of another.

\section{Experiments and Results}

This section applies the framework we developed in the previous section to several of the most popular current frontier LLMs. We will then compare how each of our selected LLMs performs relative to the GS baseline, depending on the mental health topic. Through this, we will assess the topic's impact on LLM performance. This will help us answer the following research objectives:

\textbf{RO3}. To utilise the proposed evaluation framework to assess the performance of popular frontier LLMs in mental health with verified data.

\textbf{RO4}. To study how variation in mental health topics impacts LLM nuanced conversation abilities.

\subsection{Experiment Setup}

%In order to evaluate the nuanced conversation abilities of several LLMs when dealing with mental health questions, we started by identifying the data we wish to use.

We utilise the Counsel Chat mental health dataset \cite{bertagnolli2020counsel}. Counsel Chat is a platform where individuals can ask verified and trained counsellors mental health questions. The data consists of high-quality questions and answers from domain experts. It ranges 31 topics from depression to addiction. The verified counsellors range from PhD psychologists to social workers. Additionally, we identified four of the most popular LLMs %at the time of writing 
to evaluate our model. We have provided information on the chosen LLMs in Table \ref{llmselected}. One of the primary reasons for selecting these four models is that the ones from OpenAI are state-of-the-art closed models trained on significantly more data. While trained on less data and less power, the Llama models are open-source. This means we can expect to see their wide adoption and modification for different applications in the future.

\begin{table}[h]
    \caption{Selected Large Language Models. \label{llmselected}}
    \begin{tabularx}{\textwidth}{|l|X|}
        \hline
        \textbf{LLMs} & \textbf{Brief Explanation} \\
        \hline
        GPT3.5 (Turbo) & One of OpenAI's most capable models (the most capable of the GPT3.5 range). It has been trained on data as recently as September 2021. It is optimised for chat and is highly cost-effective. Model Name: gpt-3.5-turbo.\\
        \hline
        GPT4 & OpenAI's state of the art model. It is touted to be more effective than any GPT3.5 model and highly optimised for chat and instruction. Model Name: gpt-4.\\
        \hline
        GPT3.5 Turbo 1106 & This model is effectively an improved GPT3.5 Turbo. Reproducibility, instruction following, and parallel function calling are better. Model Name: gpt-3.5-turbo-1106.\\
        \hline
        GPT4 Turbo (Preview) & This version of GPT4 is an improvement on the original and is said to be able to complete tasks fully and to a higher standard. Model Name: gpt-4-0125-preview. \\
        \hline
        Llama 2 7B & A 7 billion parameter model released by Meta. It is fully open source and performs well compared to frontier closed models in safety and helpfulness. Model Name: llama-2-7b-chat.\\
        \hline
        Llama 2 70B & Llama 2 is the 70 billion parameter version of the aforementioned Meta Llama 2 model. Another notable difference between this and the 7 billion parameter version is that this model uses an improved inference technique called Grouped Query Attention. Model Name: llama-2-70b-chat.\\
        \hline
        Mixtral 8 7B & This is a decoder only model that has open weights. It's cost to performance ratio is highly regarded. Model Name: Mixtral-8x7B-Instruct-v0.1.\\
        \hline
        Mistral 7B Instruct V1 & This is an instruction fine-tuned version of the base model. It is trained on a number of conversation datasets which are public. Model Name: Mistral-7B-Instruct-v0.1.\\
        \hline
        Mistral 7B Instruct V2 & This is an improved version of the aforementioned V1 model. Model Name: Mistral-7B-Instruct-v0.2.\\
        \hline
        
    \end{tabularx}
\end{table}

It should also be noted that we made use of a RoBERTa model \cite{Lowe}, which had been fine-tuned on emotion data, in order to conduct emotion analysis on all our relevant data. Additionally, worth mentioning is that we used Python 3 and Jupyter Notebooks for our implementations. In the following section, we provide some insight into the results we obtained following the framework and the analysis of them.

\subsection{Overall Results}
This section includes our results by applying the framework to our selected LLMs and mental health dataset. We also have various results, visualisations, and text highlighting notable and critical points.

\begin{table}[ht]
\centering
\caption{LLM Performance Across Metrics}
\resizebox{\textwidth}{!}{%
\begin{tabular}{lcccccccc}
\toprule
Model & Consistency & Sentiment Change & Intra Sentiment Change & Simplicity & Recycling & Agreeability & Active Listening \\
\midrule
GS & - & 0.7017 & 0.5003 & 86.8240 & 0.0725 & 6.6000 & 7.3294 \\
GPT3.5 & 0.5046 & \textbf{0.7063} & 0.5039 & 40.1650 & 0.0767 & 7.5804 & 8.3448 \\
GPT3.5 1106 & 0.5076 & 0.7473 & 0.5028 & 47.4323 & 0.0706 & 7.8503 & 8.4659 \\
GPT4 & 0.4737 & 0.6509 & 0.5069 & 44.7912 & \textbf{0.0732} & 7.4841 & 8.1125 \\
GPT4 Preview & \textbf{0.5229} & 0.7141 & \textbf{0.5003} & \textbf{53.9883} & 0.0696 & 7.7694 & 8.4884 \\
Llama 7 & 0.3687 & 0.6600 & 0.5022 & 33.4359 & 0.1305 & \textbf{6.5822} & \textbf{7.2287} \\
Llama 70 & 0.4113 & 0.6667 & 0.5044 & 37.8323 & 0.1224 & 6.7437 & 7.5470 \\
Mistral 7 & 0.2744 & 0.5595 & 0.5039 & 35.9182 & 0.1972 & 6.1160 & 6.7251 \\
Mistral 7 2 & 0.3959 & 0.6429 & 0.5022 & 36.0308 & 0.0977 & 6.9859 & 7.8205 \\
Mixtral 7 & 0.4248 & 0.6611 & 0.5046 & 26.8561 & 0.1118 & 6.8716 & 7.6482 \\
\bottomrule
\end{tabular}%
}
\label{llmperformance}
\end{table}

\begin{figure}[ht] 
    \centering
    \includegraphics[width=1\textwidth]
    {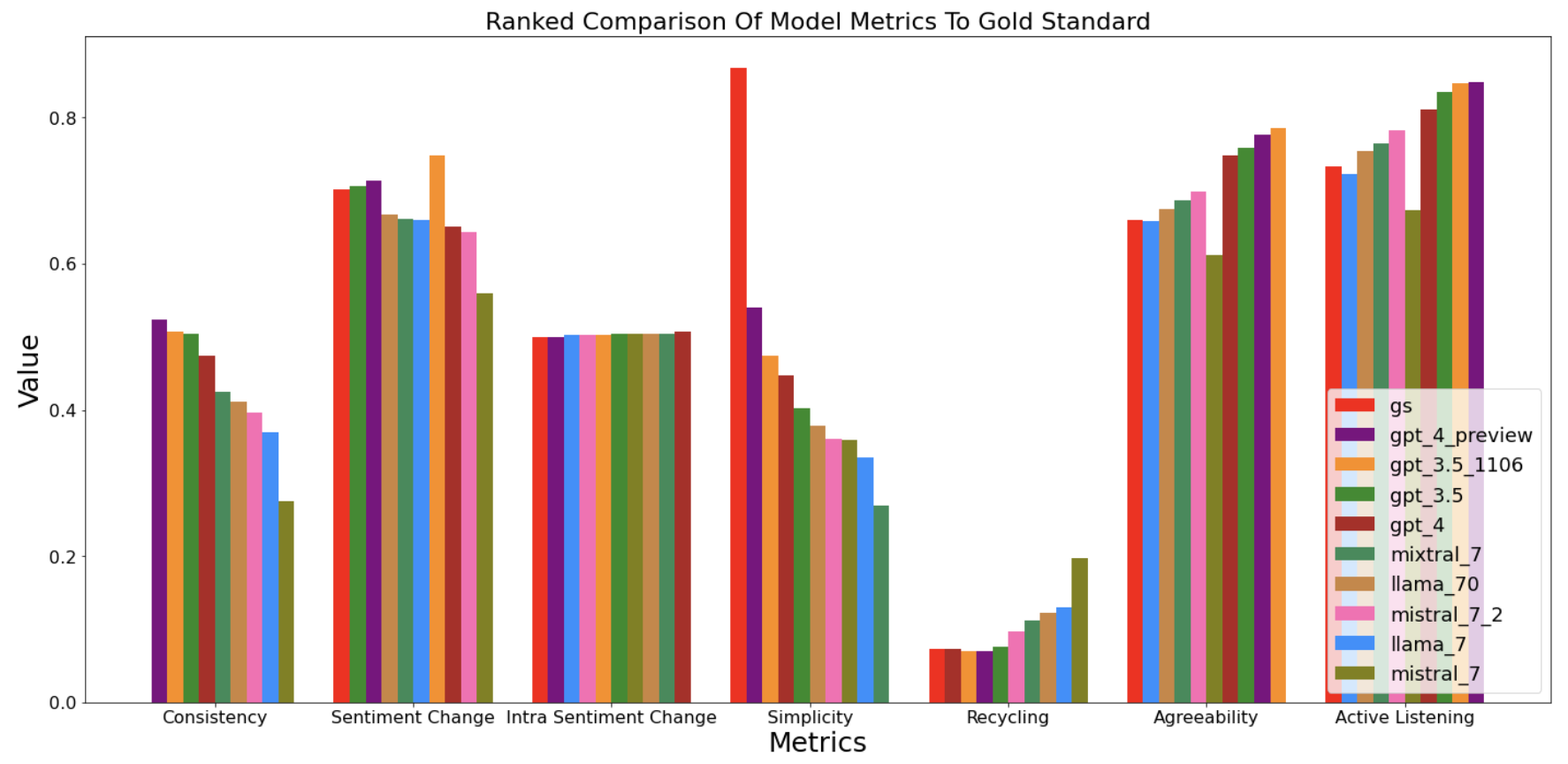} 
    \caption{Visualising Model Performance Across Metrics. \label{performance_comparison} }
\end{figure}

Table \ref{llmperformance} shows LLM performance across metrics. It is quickly worth mentioning that the GS model was the baseline verified performance. The highest performing model for our Category 1 metric, Emotion Consistency, was GPT4 Preview. Since this metric compares to the GS inherently included, the LLM with the result closest to number 1 has the highest performance. For the Category 2 metrics, performance varies across the board. For each of these metrics, we have made the LLM results closest to the GS, and hence the best performing, bold. A key highlight we can initially see from the results is that GPT4 Preview tends to have the best performance across most metrics. It's worth reaffirming that the Recycling, Agreeability, and Active Listening values will be normalised for the following sections before further statistical analysis. These results are also visualised in Figure \ref{performance_comparison}. Note that the Recycling, Agreeability, and Active Listening values will be normalised for the following sections before further statistical analysis.

When visualising the normalised results in Table \ref{pearson}, a large gap between the baseline and the rest of the models can be noticed when evaluated for Simplicity. This may indicate that the limitation we mentioned in the metric regarding Simplicity values being negatively impacted by shorter text is playing out.

\begin{table}[ht]
\centering
\caption{Pearson Correlation and P Value For Category 2 Metrics}
\footnotesize
\begin{tabular}{lcc}
\toprule
Model & Pearson Correlation & P Value \\
\midrule
GPT3.5 & 0.712032 & 0.112448 \\
GPT3.5\_1106 & 0.772984 & 0.071454 \\
GPT4 & 0.762072 & 0.078180 \\
GPT4 Preview & \textbf{0.830253} & \textbf{0.040775} \\
Llama 7 & 0.639987 & 0.171084 \\
Llama 70 & 0.691926 & 0.127745 \\
Mistral 7 & 0.632970 & 0.177345 \\
Mistral 7\_2 & 0.677051 & 0.139603 \\
Mixtral 7 & 0.566129 & 0.241529 \\
\bottomrule
\end{tabular}
\label{pearson}
\end{table}

The Pearson Correlation values across models are shown in Table \ref{pearson}; we note that GPT4 Preview has the highest correlation with the GS followed by GPT3.5\_1106. Interestingly, when applying a threshold of 0.05 for our P Value, we see that only the GPT4 Preview model has a correlation with the GS Baseline that is statistically significant. In the P-Value column, we made all values underneath the threshold bold.

\subsection{Results By Topic}

In this section, we present the results we received when analysing how correlations between the GS baseline and LLM vary based on mental health topics.

\begin{table}[ht]
\centering
\caption{Pearson Correlation Between Model and Baseline (GS) Across Topics}
\footnotesize
\begin{tabular}{lcccccccc}
\toprule
Model & Depression & Relationships & Anxiety & Intimacy & Parenting \\
\midrule
GPT3.5 & 0.727230 & 0.796242 & 0.601444 & 0.717623 & 0.723370 \\
GPT3.5\_1106 & 0.773212 & 0.843714 & 0.700000 & 0.787813 & 0.812950 \\
GPT4 & 0.795839 & 0.842861 & 0.668790 & 0.766105 & 0.762978 \\
GPT4 Preview & \textbf{0.827856} & \textbf{0.894149} & \textbf{0.761254} & \textbf{0.847404} & \textbf{0.857361} \\
Llama 7 & 0.621028 & 0.696750 & 0.563190 & 0.611425 & 0.761030 \\
Llama 70 & 0.674186 & 0.741725 & 0.573159 & 0.721064 & 0.765767 \\
Mistral 7 & 0.656170 & 0.677328 & 0.592750 & 0.621819 & 0.600908 \\
Mistral 7\_2 & 0.673585 & 0.784736 & 0.573245 & 0.670457 & 0.695712 \\
Mixtral 7 & 0.558138 & 0.674473 & 0.470480 & 0.548494 & 0.616159 \\
\bottomrule
\end{tabular}
\label{topics_performance}
\end{table}

\begin{table}[ht]
\centering
\caption{P Values Between Model and Baseline (GS) Across Topics}
\footnotesize
\begin{tabular}{lcccccccc}
\toprule
Model & Depression P-Value & Relationships P-Value & Anxiety P-Value & Intimacy P-Value & Parenting P Value \\
\midrule
GPT3.5 & 0.101458 & 0.058046 & 0.206615 & 0.108347 & 0.104202 \\
GPT3.5\_1106 & 0.071317 & \textbf{0.034729} & 0.121500 & 0.062758 & \textbf{0.049209} \\
GPT4 & 0.058267 & \textbf{0.035099} & 0.146383 & 0.075662 & 0.077611 \\
GPT4 Preview & \textbf{0.041900} & \textbf{0.016214} & 0.078695 & \textbf{0.033152} & \textbf{0.029068} \\
Llama 7 & 0.188216 & 0.123997 & 0.244532 & 0.197151 & 0.078837 \\
Llama 70 & 0.141939 & 0.091445 & 0.234406 & 0.105857 & 0.075872 \\
Mistral 7 & 0.157005 & 0.139378 & 0.215007 & 0.187488 & 0.207129 \\
Mistral 7\_2 & 0.142431 & 0.064520 & 0.234320 & 0.145004 & 0.124799 \\
Mixtral 7 & 0.249728 & 0.141704 & 0.346351 & 0.259765 & 0.192724 \\
\bottomrule
\end{tabular}
\label{zscore_performance}
\end{table}

\begin{figure}[ht] 
    \centering
    \includegraphics[width=0.8\textwidth]
    {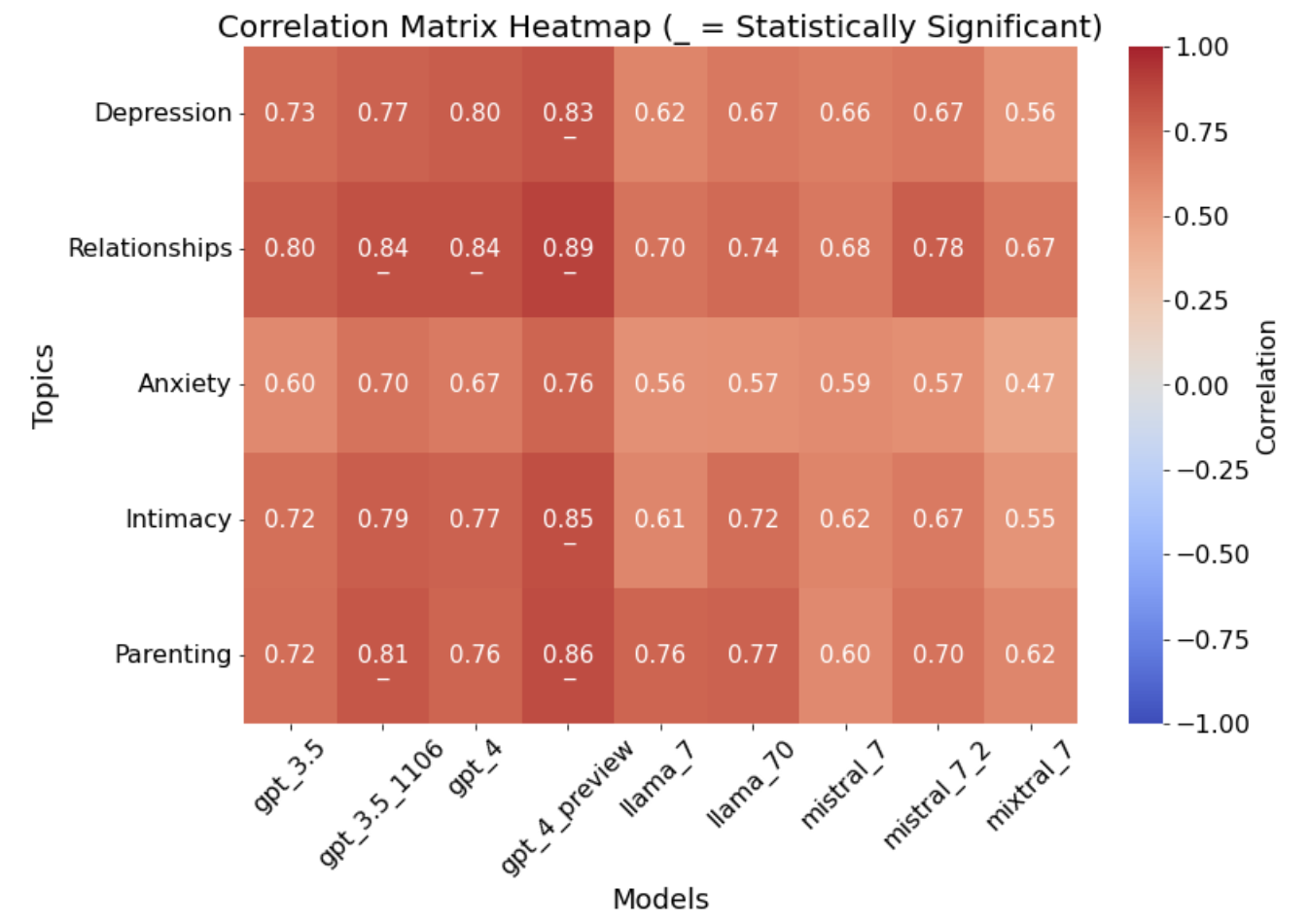} 
    \caption{Visualising Model Statistical Significance Across Topics. \label{significance_comparison} }
\end{figure}

In Table \ref{topics_performance}, we highlighted the most significant correlations from each LLMs under a specific topic. These happen to all be values from GPT4 Preview, which indicates that for all topics, GPT4 Preview has the highest correlation with the GS baseline. In Table \ref{zscore_performance}, when examining which LLMs have statistically significant correlations with the GS baseline, we see that GPT4 Preview has statistically significant correlations in all topics but Anxiety. Interestingly, none of the Models selected have statistically significant correlations with the GS in Anxiety. Additionally, under the topic of Relationships, both GPT3.5\_1106 and GPT4 also have statistically significant correlations. This is further visualised in Figure \ref{significance_comparison}.

\begin{table}[ht]
\centering
\caption{Z Scores Between Topics (Part 1)}
\footnotesize
\resizebox{\textwidth}{!}{%
\begin{tabular}{lccccccccccc}
\toprule
Model & Depression vs Relationships & Depression vs Anxiety & Depression vs Intimacy & Depression vs Parenting & Relationships vs Anxiety  \\
\midrule
GPT3.5 & -0.198997 & 0.313732 & 0.025991 & 0.010489 & 0.504841 \\
GPT3.5\_1106 & -0.223705 & 0.202222 & -0.043693 & -0.121987 & 0.420202 \\
GPT4 & -0.153315 & 0.347535 & 0.089040 & 0.098093 & 0.493020 \\
GPT4 Preview & -0.243337 & 0.204898 & -0.066943 & -0.102462 & 0.439973 \\
Llama 7 & -0.188084 & 0.134186 & 0.022811 & -0.360667 & 0.320554 \\
Llama 70 & -0.178702 & 0.241287 & -0.122342 & -0.246342 & 0.415927 \\
Mistral 7 & -0.053107 & 0.151780 & 0.083525 & 0.132840 & 0.204528 \\
Mistral 7\_2 & -0.302863 & 0.239592 & 0.007902 & -0.056788 & 0.533439 \\
Mixtral 7 & -0.275618 & 0.190975 & 0.021661 & -0.134009 & 0.461896 \\
\bottomrule
\end{tabular}
}
\label{zscoretopics_performance_part1}
\end{table}

\begin{table}[ht]
\centering
\caption{Z Scores Between Topics (Part 2)}
\footnotesize
\resizebox{\textwidth}{!}{%
\begin{tabular}{lccccccccccc}
\toprule
Model & Relationships vs Intimacy & Relationships vs Parenting & Anxiety vs Intimacy & Anxiety vs Parenting & Intimacy vs Parenting \\
\midrule
GPT3.5 & 0.224665 & 0.209364 & -0.288295 & -0.303480 & -0.015504 \\
GPT3.5\_1106 & 0.180604 & 0.102691 & -0.245300 & -0.321896 & -0.078428 \\
GPT4 & 0.241251 & 0.250151 & -0.260715 & -0.251797 & 0.009077 \\
GPT4 Preview & 0.177663 & 0.142463 & -0.270539 & -0.305081 & -0.035619 \\
Llama 7 & 0.210698 & -0.175459 & -0.111443 & -0.489642 & -0.382764 \\
Llama 70 & 0.056674 & -0.068433 & -0.361245 & -0.481051 & -0.124973 \\
Mistral 7 & 0.136498 & 0.185660 & -0.068438 & -0.019020 & 0.049431 \\
Mistral 7\_2 & 0.310578 & 0.247165 & -0.231785 & -0.295496 & -0.064684 \\
Mixtral 7 & 0.296923 & 0.142707 & -0.169443 & -0.323422 & -0.155581 \\
\bottomrule
\end{tabular}
}
\label{zscoretopics_performance_part2}
\end{table}

In Table \ref{zscoretopics_performance_part1} and \ref{zscoretopics_performance_part2}, we computed Steiger's Z test to assess if, under certain mental health topics, the LLMs performed closer to the GS baseline than in others. 
Results where the absolute value of the Z score is more significant than 1.96 indicate statistical significance. It can be observed that there are no statistically significant differences in performance across topics. 

\begin{figure}[ht] 
    \centering
    \includegraphics[width=0.8\textwidth]
    {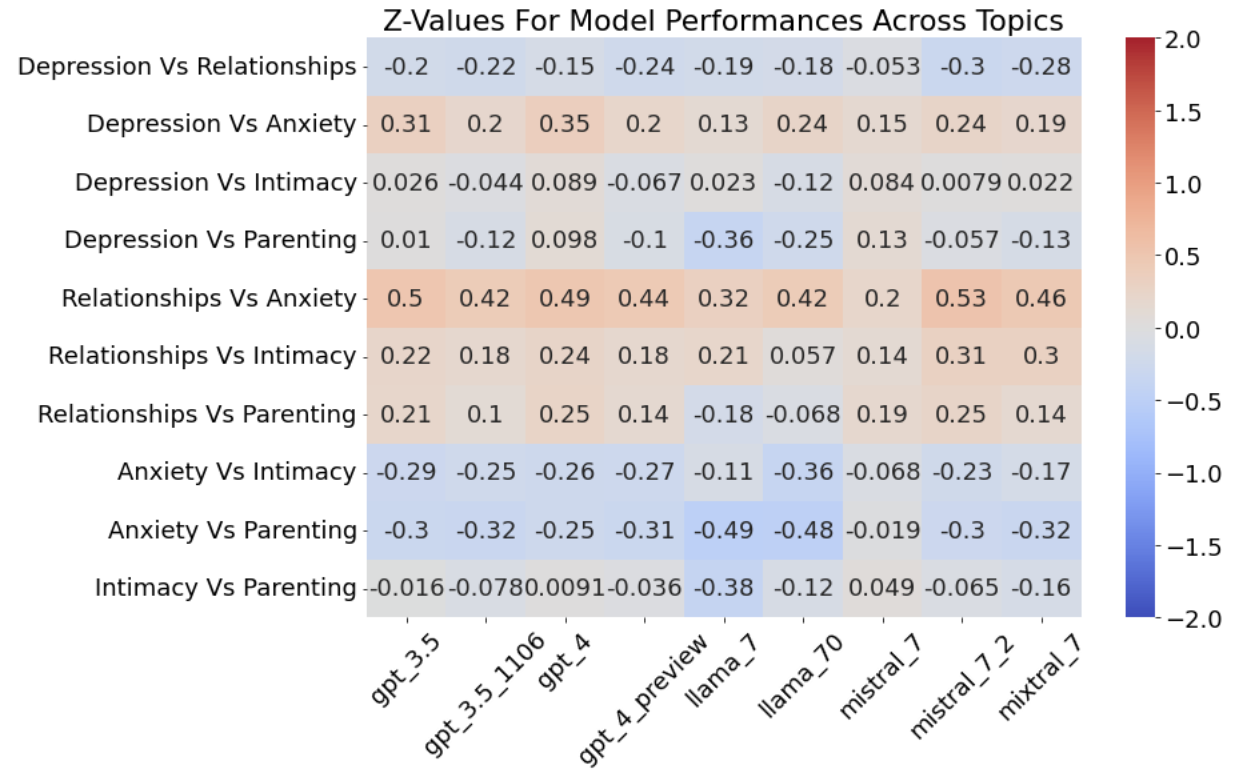} 
    \caption{Visualising Z Scores. \label{figure_heatmap} }
\end{figure}

In Figure \ref{figure_heatmap}, the Z Scores are represented as a heat map for the results shown in Table \ref{zscoretopics_performance_part1} and \ref{zscoretopics_performance_part2}. This helps provide an intuitive way to grasp the information presented in Table \ref{zscoretopics_performance_part1} and \ref{zscoretopics_performance_part2}. The more intense the colour of a cell (either dark red or dark blue), the larger the Z score value. The cells where the absolute value of the Z score is greater than 1.96 are statistically significant. 

\section{Discussion} 

In this section, we start with an interpretation of the results and a discussion of some of their strengths and weaknesses. We then explain how our work has been methodically conducted to answer our research objectives and overarching goal. We then take some time to cover the specific contributions of our work as well as the importance of them in a broader context. To do this, we will also discuss how the work fits into the existing literature. Finally, we will present some of the limitations of the work in general.

%As mentioned, we will begin by interpreting the results we produced in the previous section. 
For our Category 1 metric, Emotion Consistency, GPT4 Preview had superior performance to other LLMs. This means it was able to produce similar emotions to verified mental health therapists most closely when answering questions. Across our Category 2 metrics, GPT4 Preview had the highest correlation with our verified baseline. Additionally, it was the only statistically significant correlation. This is informative for providing a holistic sense of each LLM's relative directional similarity to verified mental health counsellors. One caveat of our use of the Pearson correlation is that it does not provide information regarding the absolute distance between the GS and LLM for specific metrics. When evaluating specific mental health topics, GPT4 Preview had the highest correlation with the GS baseline in every topic. Additionally, all of its correlations, excluding that for Anxiety, were shown to be statistically significant. GPT3.5\_1106 and GPT4 were both shown to be statistically significant in Relationships. Interestingly, none of the selected Models had statistically significant correlations in Anxiety. This is essential information that can help provide guidance regarding which LLMs to use when dealing with specific domains in mental health. This may be exceedingly useful when planning to use GPT3.5\_1106 for a mental health chatbot. Through our results, they may be able to make the connection that when dealing with questions regarding Depression, GTP3.5\_1106 is not able to perform similarly enough to verify therapists in nuanced language abilities. By proactively selecting a better-suited Model for the situation, such as GPT4 Preview, they can improve the mental health chatbot performance. In turn, this may help users navigate challenging psychological scenarios more effectively. This reaffirms the importance of understanding which topics LLMs are operating in, as it can significantly impact the nuanced conversation abilities of the LLMs.

We also take the time to briefly highlight some notable differences between the tested models in Table \ref{brief_model_comparison}. This includes insight from our results as well as general characteristics worth reaffirming. The highlighted model differences are highly informative for readers to understand performance and characteristic variation. 

\begin{table}[ht]
    \caption{Model Comparison \label{brief_model_comparison}}
    \begin{tabularx}{\textwidth}{|l|X|}
        \hline
        \textbf{Model} & \textbf{Analysis} \\
        \hline
        GPT3.5 (Turbo) & GPT3.5 demonstrated a Sentiment Change performance that most closely matched the baseline. Despite this, its Pearson correlations were neither the highest nor statistically significant. The model boasts a context window of 4,096 Tokens, the least of the following GPT models tested.\\
        \hline
        GPT4 & GPT4 performed the most similarly to the GS verified baseline when being measured for Recycling. Additionally, it was shown to have a statistically significant correlation with the baseline when dealing with questions related to Relationships. It has a context window of 8,192 tokens. \\
        \hline
        GPT3.5 Turbo 1106 & GPT3.5 Turbo 1106 did not perform in any of the metrics closest to the GS baseline. However, its Pearson Correlation with the GS baseline was the second highest. It is worth mentioning that based on the P-Value, the overall correlation appears not statistically significant. When we look at the specific topics, the GPT3.5 1106 Pearson Correlation, specifically for Relationships and Parenting, appears to be statistically significant. It has a context window of 16,385 tokens.\\
        \hline
        GPT4 Preview & The GPT4 Preview LLM boosted the most notable results. It had a performance under Intra Sentiment Change and Simplicity, which was the closest to the GS baseline. It had the highest overall Pearson Correlation and the only statistically significant one. In addition, it had the highest correlation in all topics and was statistically significant in all of them, excluding Anxiety. The tested GPT models have the largest context window of 128,000 Tokens. \\
        \hline
        Llama 2 7B & The 7B parameter version of Llama 2 has a performance under Agreeability and Active Listening, which most closely matches that of the GS baseline. Despite that, it doesn't have an overall statistically significant correlation, nor does it for any of the specific topics. Llama 2 has a context window of 4,096 Tokens.\\
        \hline
        Llama 2 70B & The larger 70B parameter version of Llama 2 trails the 7B version with the second-best Agreeability and Active Listening performance. However, when it comes to the Correlations, none of them appear to be statistically significant. It also has a context window of 4,096 Tokens.\\
        \hline
        Mixtral 8 7B & Mixtral has around 12B active parameters and requires more RAM and GPUs to fine-tune than Mistral 7B. Mixtral's performance is average in regard to distance from the GS baseline. It has no statistically significant correlations, overall and for specific topics. Mistral has a context window of 32k tokens.\\
        \hline
        Mistral 7B Instruct V1 & This is a fine-tuned version of Mistral designed to perform exceptionally well at following instructions. Mistral 7B V1's performance is poor compared to the other Models and produces results under metrics like Consistency and Sentiment Change, which are the farthest from the GS baseline. \\
        \hline
        Mistral 7B Instruct V2 & This is considered an improved instruction fine-tuned version of the above Mistral 7B Instruct V1. This is backed up by the fact that it performs more closely to the GS baseline in all metrics. However, its overall and topic-specific correlations are both not statistically significant.\\
        \hline
        
    \end{tabularx}
\end{table}

To provide a comprehensive view to readers regarding effective LLM selection, we will spend time mentioning factors worth considering in detail before any LLM integration. Even if these factors are not the focus of this paper, they are worth considering.\\
\textit{Additional Factors To Consider For Users}: 
\begin{itemize}
    \item \textbf{Utility}: The quality of the content suggested by the LLMs is essential. This is something that can manifest itself through high-quality, relevant training data. Because of this, it is necessary to examine LLM performance on functional benchmarks if possible. It is important to note that this is also how up-to-date the training data is. For example, the GPT models selected leverage data up to September 2021. 
    \item \textbf{Responsiveness}: Given the potential sensitive domains in which these LLMs can be deployed, the user experience regarding response time is significant. Having to wait for long periods for a response may frustrate users. Given that LLM response time is based on output token quantity, we can expect models like GPT4 to take longer than  GPT3.
    \item \textbf{User Interface}: The general user interface is another aspect worth considering. While hard to define due to its nuanced nature, the aesthetic appearance of the interface may make it easier or harder for users to have meaningful engagement with them.
    
\end{itemize}
\textit{Additional Factors To Consider For Vendors}: 
\begin{itemize}
    \item \textbf{Cost}: A significant factor to consider is the cost. For example, at the time of writing, the cost of a GPT4 output token is around 40 times more expensive than that of a GPT3.5 Turbo output token. Depending on budget constraints and the system's potential impact, this is worth considering in detail.
    \item \textbf{Integration}: Integration with existing systems is essential for vendors to consider. There may be cases where the API calls can't be made due to legacy technology systems. However, given the popularity of these specific models, they may not be problematic for the task at hand.
    \item \textbf{Scalability}: This is another factor worth considering. Given that some models use more resources than others, such as Mixtral 8 7B over Mistral 7B, we must consider the potential use scale. If we want this to be deployed to a small batch and want high performance, we may choose a more resource-intensive model such as Mixtral or an expensive one like GPT4. If we want to deploy this at scale and reduce costs, we may select GPT3.5 Turbo or Mistral 7B Insturct V2.
    
\end{itemize}

We now discuss how we designed our methodology and provide a justification for the approach used. This will enable us to explain how our work helped us cover our research objectives and answer our overarching research goal. 

Firstly, we developed a procedural and translatable framework to assist researchers in evaluating LLMs for nuanced conversation abilities. This helped us answer \textbf{RO1} (To propose a novel evaluation framework to analyse the nuanced conversation abilities of Large Language Models.) Within the proposed framework, we developed several metrics that would help provide quantifiable insights into LLM's nuanced conversational abilities. These metrics were justified and developed by leveraging existing counselling literature. This helped us answer \textbf{RO2} (To develop several quantitative metrics, which vary from measuring affective content to emulating conversation strategies from counselling literature.) We then applied our proposed evaluation framework to several of the most popular frontier LLMs and paired this with a verified mental health dataset. We did this to assess how these LLMs compare to a verified baseline in answering mental health questions. This helped us reach the following research objective \textbf{RO3} (To utilise the proposed evaluation framework to assess the performance of popular frontier LLMs in mental health with verified data.) We then conducted an extended analysis to study how different mental health topics impacted LLM performance. This helped us answer the following research goal \textbf{RO4} (To study how variation in mental health topics impacts LLM nuanced conversation abilities.) 

Through studying and tailoring our method to the research objectives, we were able to create a study that helped effectively answer our overarching research goal: To conduct a study into the nuanced conversation abilities of popular frontier LLMs when dealing with mental health questions, which can be leveraged by researchers for application to adjacent domains.

%We will now take some time to consider our specific contributions, their implications, and how they fit into existing literature. Following the completion of our research objectives, we can establish our contributions to the field as follows:

%\begin{itemize}
%    \item A novel framework researchers can use to guide their work in analysing the nuanced conversational abilities of LLMs.
%    \item A series of quantitative metrics to analyse conversational performance, which have been developed from psychotherapy conversation analysis literature.
%    \item The evaluation of the nuanced conversation abilities of four leading frontier LLMs, when dealing with verified mental health questions.
%    \item An analysis of how mental health topic impacts conversation performance of leading frontier LLMs.
    
%\end{itemize}

The broader implications of our contributions range depending on the stakeholder. For researchers attempting to apply this work to their field, our framework and metrics provides a systematic way to analyse the nuanced conversation abilities of LLMs in a variety of contexts. This means they will be able to make smarter decisions about which LLMs to use in which situations. The importance of this is that people will be able to evaluate LLMs effectively and then decide to employ them in situations where they can safely help people. The downstream effects of this are that people interacting with LLMs will be able to have better conversations that, depending on context, could hugely impact their lives. An example of this being a first-responder emergency LLM that is able to effectively interact with and help users. 

Due to the novel nature of the topic, there do not exist exactly relevant LLM mental health baselines to compare the contributions of our paper to relevant works. However, we mentioned related existing works that can help provide contextual information regarding our work. There are very limited frameworks that evaluate empathy over conversations \cite{Amjad2023}. However, besides just focusing on empathy, this framework is not specific to the LLMs. %We have also seen papers put forth 
Another work \cite{Lin2023} carried out evaluation for LLMs that focus specifically on analysing LLM dialogues. However, this work does not provide an effective framework which can be applied by others, in addition, the proposed metrics are not nuanced (i.e., they measure aspects like grammar). Another work studies LLMs in the specific field of mental health \cite{Chen2023}, however, this work does not focus on evaluating nuanced conversation abilities (i.e., focuses more on measuring engagement with the chatbot). Finally, we have also seen papers that focus on providing some form of applied metrics (in this case to understand empathy) to the field of mental health \cite{Sharma2020}. However, the use of unverified data makes it difficult for us to compare to. Additionally, while we focus on analysing some of the most popular existing frontier LLMs, this paper uses a custom RoBERTa model. We believe our work is closely related to and further enhances this framework. Firstly, it is a transferable framework for researchers to use to study the nuanced conversation abilities of LLMs, with a suite of quantitative metrics. Secondly, it is a study of a number of the most popular existing LLMs to evaluate how they perform when responding to the verified mental health questions.

A potential limitation is our use of Pearson correlation as a similarity measure. This is something researchers should bear in mind while crafting their implementations of the framework. We proposed it as it provides useful information regarding the directional similarity of LLM performance to the GS baseline. However, it does not provide information regarding the absolute distance from the GS baseline. It is also important to acknowledge that due to privacy issues, genuine verified healthcare data is hard to access, hence, we were only able to use one dataset. %which was not as detailed as we would have liked. 
%Any inherent issues with this dataset would have bled through to our results. The fact that the data is a singular question and answer also provides limitations due to the fact that many of the conversation analysis metrics we implemented were originally designed for conversations that consist of multiple turns. 
Additionally, the counseling techniques which we implemented into our metrics are generally used to evaluate conversations, which occurred in person. %Seeing as the data we used was for that over the internet, there may be a mismatch. 
In order to provide as much substance as possible to our topic analysis, we analysed the topics which existed in the highest quantity within our dataset. %However, it is important to acknowledge that each specific topic has no more than 50 entries. 
%This lack of substantial amounts of data may mean the topic analysis is weaker than we would like. 
Another limitation is that some of the questions or answers inside the utilised dataset are not that long, hence our metrics are based on analysing properties of answers would have less to work with. This is a problem for metrics such as Simplicity, which require a certain response length. It is likely because of this that we see large variation in GS and LLM Simplicity values in our results. For Active Listening and Agreeability, we use GPT3.5 Turbo to produce the results. A potential limitation could be if GPT3.5 is more likely to evaluate its own generated text better than that of GPT4 or one of the Llama models. %Something else to consider may be the impact of the fact that the LLMs were trained on the dataset. Additionally, any bias in GPT3.5's training data may impact results for Active Listening and Agreeability values.  

\section{Conclusion and Future Work} 
%\subsection{Conclusion}

\paragraph{Conclusion}In this paper, we have provided a framework and series of quantitative metrics that enable the evaluation of LLM's nuanced conversation abilities. While other papers have put forward frameworks to measure aspects such as empathy in dialogue \cite{Amjad2023}, ours is the first to incorporate LLMs into itself while providing a more comprehensive suite of applied metrics that measure more than just empathetic abilities. In addition, we have applied the framework to the mental health field to evaluate the abilities of several of the most famous frontier LLMs at answering mental health questions. While there do exist studies that have measured empathy expressed by LLMs in mental health \cite{Sharma2020}, our use of verified data and frontier popular LLMs help provide practical and insightful results that span past just empathy. In our evaluation, we have identified which LLMs can most accurately match the conversational abilities of a verified therapist baseline directionally. Additionally, we have investigated variance in LLM performance across mental health topics.

Our proposed framework and metrics enable researchers to evaluate the nuanced conversation abilities of LLMs across relevant domains. This will lead to more intelligent and thoughtful implementation, which will lead to LLMs being able to positively impact the lives of more people. Additionally, our specific analysis of LLMs in the mental health field will provide insight in more ways than one. Individuals who may be looking for help with their mental health questions can use our findings to help guide which LLMs may be able to best respond similarly to a verified therapist and for specific mental health topics. For example, someone suffering from Depression may be able to use our results to help guide their decision to talk to GPT4 Preview rather than GPT3.5. Researchers looking to build dedicated mental health chatbots can use our findings to guide their development and end up with a better product.

However, our work does contain certain limitations. Some limitations include the fact that our metrics are adapted from conversational counseling literature, which may make some of them less suited for analyzing question and answer-based text. Additionally, the use of GPT3.5 to explore some of our data may also introduce bias from the training data of that specific model.

%\subsection{Future Work}
\paragraph{Future Work}
There are many ways in which this body of work can be extended in the future. Firstly, regarding the framework, additional metrics can be created to help encapsulate other conversation facets. An exciting avenue to explore for this can be incorporating language and culture-specific metrics. This is because particular languages may have certain aspects which are unique to them. For example, some may use a lot of slang, some may be straightforward, and some may use more explicit vocabulary. When it comes to the application of the framework, while we did so in the domain of mental health, there are several other avenues that it can be applied to. Future work can evaluate LLM performance with our framework in avenues such as first-responder emergency calls, primary care questions for general practitioners in healthcare, and financial advice questions for investment managers. We can generalize recommended future directions for applying the framework to any domain where information presentation impacts its quality.

%\section*{Acknowledgments}
%This was was supported in part by......

%Bibliography
\bibliographystyle{unsrt}  
\bibliography{thesis/references}

\begin{thebibliography}{10}

\bibitem{Reddy2023}
Sandeep Reddy.
\newblock Evaluating large language models for use in healthcare: A framework for translational value assessment.
\newblock {\em Informatics in Medicine Unlocked}, 41:101304, 2023.

\bibitem{Amjad2023}
Bushra Amjad, Muhammad Zeeshan, and Mirza~Omer Beg.
\newblock Emp-eval: A framework for measuring empathy in open domain dialogues, 2023.

\bibitem{Sharma2020}
Ashish Sharma, Adam~S. Miner, David~C. Atkins, and Tim Althoff.
\newblock A computational approach to understanding empathy expressed in text-based mental health support, 2020.

\bibitem{Vaswani2017}
Ashish Vaswani, Noam Shazeer, Niki Parmar, Jakob Uszkoreit, Llion Jones, Aidan~N. Gomez, Lukasz Kaiser, and Illia Polosukhin.
\newblock Attention is all you need, 2023.

\bibitem{Devlin2019BERTPO}
Jacob Devlin, Ming-Wei Chang, Kenton Lee, and Kristina Toutanova.
\newblock Bert: Pre-training of deep bidirectional transformers for language understanding.
\newblock In {\em North American Chapter of the Association for Computational Linguistics}, 2019.

\bibitem{Liu2019RoBERTaAR}
Yinhan Liu, Myle Ott, Naman Goyal, Jingfei Du, Mandar Joshi, Danqi Chen, Omer Levy, Mike Lewis, Luke Zettlemoyer, and Veselin Stoyanov.
\newblock Roberta: A robustly optimized bert pretraining approach.
\newblock {\em ArXiv}, abs/1907.11692, 2019.

\bibitem{Ji2021}
Shaoxiong Ji, Tianlin Zhang, Luna Ansari, Jie Fu, Prayag Tiwari, and Erik Cambria.
\newblock Mentalbert: Publicly available pretrained language models for mental healthcare, 2021.

\bibitem{Lowe}
Sam Lowe.
\newblock roberta-base-go\_emotions llm model.
\newblock \url{https://huggingface.co/SamLowe/roberta-base-go_emotions}, 2022.

\bibitem{10.5555/3495724.3495883}
Tom~B. Brown, Benjamin Mann, Nick Ryder, Melanie Subbiah, Jared Kaplan, Prafulla Dhariwal, Arvind Neelakantan, Pranav Shyam, Girish Sastry, Amanda Askell, Sandhini Agarwal, Ariel Herbert-Voss, Gretchen Krueger, Tom Henighan, Rewon Child, Aditya Ramesh, Daniel~M. Ziegler, Jeffrey Wu, Clemens Winter, Christopher Hesse, Mark Chen, Eric Sigler, Mateusz Litwin, Scott Gray, Benjamin Chess, Jack Clark, Christopher Berner, Sam McCandlish, Alec Radford, Ilya Sutskever, and Dario Amodei.
\newblock Language models are few-shot learners.
\newblock In {\em Proceedings of the 34th International Conference on Neural Information Processing Systems}, NIPS'20, Red Hook, NY, USA, 2020. Curran Associates Inc.

\bibitem{Touvron2023}
Hugo Touvron, Louis Martin, Kevin Stone, Peter Albert, Amjad Almahairi, Yasmine Babaei, Nikolay Bashlykov, Soumya Batra, Prajjwal Bhargava, Shruti Bhosale, Dan Bikel, Lukas Blecher, Cristian~Canton Ferrer, Moya Chen, Guillem Cucurull, David Esiobu, Jude Fernandes, Jeremy Fu, Wenyin Fu, Brian Fuller, Cynthia Gao, Vedanuj Goswami, Naman Goyal, Anthony Hartshorn, Saghar Hosseini, Rui Hou, Hakan Inan, Marcin Kardas, Viktor Kerkez, Madian Khabsa, Isabel Kloumann, Artem Korenev, Punit~Singh Koura, Marie-Anne Lachaux, Thibaut Lavril, Jenya Lee, Diana Liskovich, Yinghai Lu, Yuning Mao, Xavier Martinet, Todor Mihaylov, Pushkar Mishra, Igor Molybog, Yixin Nie, Andrew Poulton, Jeremy Reizenstein, Rashi Rungta, Kalyan Saladi, Alan Schelten, Ruan Silva, Eric~Michael Smith, Ranjan Subramanian, Xiaoqing~Ellen Tan, Binh Tang, Ross Taylor, Adina Williams, Jian~Xiang Kuan, Puxin Xu, Zheng Yan, Iliyan Zarov, Yuchen Zhang, Angela Fan, Melanie Kambadur, Sharan Narang, Aurelien Rodriguez, Robert Stojnic, Sergey Edunov, and Thomas
  Scialom.
\newblock Llama 2: Open foundation and fine-tuned chat models, 2023.

\bibitem{Luzniak2023}
Karolina Luzniak.
\newblock 6 main differences between llama 2, gpt\-3.5 \& gpt\-4, 2023.
\newblock online article.

\bibitem{Dillet2023}
Romain Dillet.
\newblock Mistral ai, a paris-based openai rival, closed its \$415 million funding round.
\newblock {\em TechCrunch}, Dec 2023.

\bibitem{jiang2023mistral}
Albert~Q. Jiang, Alexandre Sablayrolles, Arthur Mensch, Chris Bamford, Devendra~Singh Chaplot, Diego de~las Casas, Florian Bressand, Gianna Lengyel, Guillaume Lample, Lucile Saulnier, Lélio~Renard Lavaud, Marie-Anne Lachaux, Pierre Stock, Teven~Le Scao, Thibaut Lavril, Thomas Wang, Timothée Lacroix, and William~El Sayed.
\newblock Mistral 7b, 2023.

\bibitem{Kessler2005}
Ronald~C. Kessler, Patricia Berglund, Olga Demler, Robert Jin, Kathleen~R. Merikangas, and Ellen~E. Walters.
\newblock Lifetime prevalence and age-of-onset distributions of dsm-iv disorders in the national comorbidity survey replication.
\newblock {\em Archives of General Psychiatry}, 62:593, 6 2005.

\bibitem{Lambert2002}
Michael~J. Lambert and Allen~E. Bergin.
\newblock The effectiveness of psychotherapy.
\newblock In Michel Hersen and William Sledge, editors, {\em Encyclopedia of psychotherapy}, volume~1, pages 709--714. Elsevier Science, USA, 2002.

\bibitem{Hellebuyck2019}
Michele Hellebuyck, Madeline Halpern, Theresa Nguyen, and Danielle Fritze.
\newblock The state of mental health in america, 2019.
\newblock Report.

\bibitem{Nelson2020}
Benjamin~W. Nelson, Adam Pettitt, Jessica~E. Flannery, and Nicholas~B. Allen.
\newblock Rapid assessment of psychological and epidemiological correlates of covid-19 concern, financial strain, and health-related behavior change in a large online sample.
\newblock {\em PLOS ONE}, 15:e0241990, 11 2020.

\bibitem{UpworkStudy}
{Upwork Press Releases}.
\newblock Upwork study finds 22\% of american workforce will be remote by 2025.
\newblock \url{https://www.upwork.com/press/releases/upwork-study-finds-22-of-american-workforce-will-be-remote-by-2025}, 2020.

\bibitem{Ralston2019}
Allura~L. Ralston, Arthur~R. Andrews, and Debra~A. Hope.
\newblock Fulfilling the promise of mental health technology to reduce public health disparities: Review and research agenda.
\newblock {\em Clinical Psychology: Science and Practice}, 26, 3 2019.

\bibitem{Weiste2013}
Elina Weiste and Anssi Peräkylä.
\newblock A comparative conversation analytic study of formulations in psychoanalysis and cognitive psychotherapy.
\newblock {\em Research on Language \& Social Interaction}, 46:299--321, 10 2013.

\bibitem{Vehviläinen2008}
Sanna Vehviläinen, Anssi Peräkylä, Charles Antaki, and Ivan Leudar.
\newblock {\em A review of conversational practices in psychotherapy}, pages 188--197.
\newblock Cambridge University Press, 4 2008.

\bibitem{Liu2023}
Hanmeng Liu, Ruoxi Ning, Zhiyang Teng, Jian Liu, Qiji Zhou, and Yue Zhang.
\newblock Evaluating the logical reasoning ability of chatgpt and gpt-4, 2023.

\bibitem{Wang2023}
Boxin Wang, Weixin Chen, Hengzhi Pei, Chulin Xie, Mintong Kang, Chenhui Zhang, Chejian Xu, Zidi Xiong, Ritik Dutta, Rylan Schaeffer, Sang~T. Truong, Simran Arora, Mantas Mazeika, Dan Hendrycks, Zinan Lin, Yu~Cheng, Sanmi Koyejo, Dawn Song, and Bo~Li.
\newblock Decodingtrust: A comprehensive assessment of trustworthiness in gpt models, 2024.

\bibitem{Bang2023}
Yejin Bang, Samuel Cahyawijaya, Nayeon Lee, Wenliang Dai, Dan Su, Bryan Wilie, Holy Lovenia, Ziwei Ji, Tiezheng Yu, Willy Chung, Quyet~V. Do, Yan Xu, and Pascale Fung.
\newblock A multitask, multilingual, multimodal evaluation of chatgpt on reasoning, hallucination, and interactivity, 2023.

\bibitem{Lin2004ROUGEAP}
Chin-Yew Lin.
\newblock Rouge: A package for automatic evaluation of summaries.
\newblock In {\em Annual Meeting of the Association for Computational Linguistics}, 2004.

\bibitem{Lin2023}
Yen-Ting Lin and Yun-Nung Chen.
\newblock Llm-eval: Unified multi-dimensional automatic evaluation for open-domain conversations with large language models, 2023.

\bibitem{Zheng2023}
Lianmin Zheng, Wei-Lin Chiang, Ying Sheng, Tianle Li, Siyuan Zhuang, Zhanghao Wu, Yonghao Zhuang, Zhuohan Li, Zi~Lin, Eric.~P Xing, Joseph~E. Gonzalez, Ion Stoica, and Hao Zhang.
\newblock Lmsys-chat-1m: A large-scale real-world llm conversation dataset, 2023.

\bibitem{Chang2023}
Yupeng Chang, Xu~Wang, Jindong Wang, Yuan Wu, Linyi Yang, Kaijie Zhu, Hao Chen, Xiaoyuan Yi, Cunxiang Wang, Yidong Wang, Wei Ye, Yue Zhang, Yi~Chang, Philip~S. Yu, Qiang Yang, and Xing Xie.
\newblock A survey on evaluation of large language models, 2023.

\bibitem{Huang2023}
Jen tse Huang, Man~Ho Lam, Eric~John Li, Shujie Ren, Wenxuan Wang, Wenxiang Jiao, Zhaopeng Tu, and Michael~R. Lyu.
\newblock Emotionally numb or empathetic? evaluating how llms feel using emotionbench, 2024.

\bibitem{Heinz2023}
Michael~V. Heinz, Sukanya Bhattacharya, Brianna Trudeau, Rachel Quist, Seo~Ho Song, Camilla~M. Lee, and Nicholas~C. Jacobson.
\newblock Testing domain knowledge and risk of bias of a large-scale general artificial intelligence model in mental health.
\newblock {\em DIGITAL HEALTH}, 9:205520762311704, 1 2023.

\bibitem{Nguyen2022}
Thong Nguyen, Andrew Yates, Ayah Zirikly, Bart Desmet, and Arman Cohan.
\newblock Improving the generalizability of depression detection by leveraging clinical questionnaires, 2022.

\bibitem{Xu2023}
Xuhai Xu, Bingsheng Yao, Yuanzhe Dong, Saadia Gabriel, Hong Yu, James Hendler, Marzyeh Ghassemi, Anind~K. Dey, and Dakuo Wang.
\newblock Mental-llm: Leveraging large language models for mental health prediction via online text data, 2024.

\bibitem{Mansoori2022}
Madiha Mansoori, Hrishil Maliwal, Sharvil Kotian, Hersh Kenkre, Ishani Saha, and Payal Mishra.
\newblock A systematic survey on computational agents for mental health aid.
\newblock In {\em 2022 IEEE 7th International conference for Convergence in Technology (I2CT)}, pages 1--7, 2022.

\bibitem{Chen2023}
Siyuan Chen, Mengyue Wu, Kenny~Q. Zhu, Kunyao Lan, Zhiling Zhang, and Lyuchun Cui.
\newblock Llm-empowered chatbots for psychiatrist and patient simulation: Application and evaluation, 2023.

\bibitem{Papineni2001}
Kishore Papineni, Salim Roukos, Todd Ward, and Wei-Jing Zhu.
\newblock {B}leu: a method for automatic evaluation of machine translation.
\newblock In Pierre Isabelle, Eugene Charniak, and Dekang Lin, editors, {\em Proceedings of the 40th Annual Meeting of the Association for Computational Linguistics}, pages 311--318, Philadelphia, Pennsylvania, USA, July 2002. Association for Computational Linguistics.

\bibitem{Reddy2021}
Sandeep Reddy, Wendy Rogers, Ville-Petteri Makinen, Enrico Coiera, Pieta Brown, Markus Wenzel, Eva Weicken, Saba Ansari, Piyush Mathur, Aaron Casey, and Blair Kelly.
\newblock Evaluation framework to guide implementation of ai systems into healthcare settings.
\newblock {\em BMJ Health \& Care Informatics}, 28:e100444, 10 2021.

\bibitem{Finlay2015}
Linda Finlay.
\newblock {\em Relational Integrative Psychotherapy}, volume Website Edition.
\newblock Wiley, 9 2015.

\bibitem{Nelson-Jones2005}
Richard Nelson-Jones.
\newblock Practical counselling and helping skills: Text and activities for the lifeskills counselling model.
\newblock {\em Sage Publications Ltd.}, 5th Edition, 2005.

\bibitem{Joshi2021}
Preetam Joshi.
\newblock Rbo v/s kendall tau to compare ranked lists of items, 1 2021.

\bibitem{loria2018textblob}
Steven Loria.
\newblock textblob documentation.
\newblock {\em Release 0.15}, 2, 2018.

\bibitem{Schegloff2007}
Emanuel~A. Schegloff.
\newblock {\em Sequence Organization in Interaction}.
\newblock Cambridge University Press, 1 2007.

\bibitem{Syzdek2020}
Brian~M. Syzdek.
\newblock Client and therapist psychotherapy sentiment interaction throughout therapy.
\newblock {\em Psychological Studies}, 65:520--530, 12 2020.

\bibitem{ReadabilitySoftware}
Carmine DiMAscio.
\newblock {Python Readability Metrics}.
\newblock \url{https://pypi.org/project/py-readability-metrics/}, 2020.

\bibitem{Madill2001}
Anna Madill, Sue Widdicombe, and Michael Barkham.
\newblock The potential of conversation analysis for psychotherapy research.
\newblock {\em The Counseling Psychologist}, 29:413--434, 5 2001.

\bibitem{Holmqvist2016}
Rolf Holmqvist, Björn Philips, and John Mellor-Clark.
\newblock Client and therapist agreement about the client's problems—associations with treatment alliance and outcome.
\newblock {\em Psychotherapy Research}, 26:399--409, 7 2016.

\bibitem{Hirsh2012}
Jacob~B. Hirsh, Lena~C. Quilty, R.~Michael Bagby, and Shelley~F. McMain.
\newblock The relationship between agreeableness and the development of the working alliance in patients with borderline personality disorder.
\newblock {\em Journal of Personality Disorders}, 26:616--627, 8 2012.

\bibitem{Xiao2020}
Ziang Xiao, Michelle~X. Zhou, Wenxi Chen, Huahai Yang, and Changyan Chi.
\newblock If i hear you correctly: Building and evaluating interview chatbots with active listening skills.
\newblock In {\em Proceedings of the 2020 CHI Conference on Human Factors in Computing Systems}, CHI ’20. ACM, April 2020.

\bibitem{bertagnolli2020counsel}
Nicolas Bertagnolli.
\newblock Counsel chat: Bootstrapping high-quality therapy data, 2020.

\end{thebibliography}

\end{document}